\documentclass{article}

\title{Domain Generalization under Sampling Pattern Shifts in Irregular Time Series}

\author{Changhun Kim$^{1,2}$ \hquad Joohyung Lee$^{2}$ \hquad Kwanhyung Lee$^{2,3}$ \hquad Donghwee Yoon$^{2}$ \\ \textbf{Grigorios Chrysos}$^{1}$ \hquad \textbf{Eunho Yang}$^{2,3}$ \\
$^1$University of Wisconsin--Madison \hquad $^2$AITRICS \hquad $^3$KAIST \\
\texttt{\{changhun.kim, chrysos\}@wisc.edu} \\
\texttt{\{soteria, kwanlee9209, dhyoon, eunhoy\}@aitrics.com} 
}

\usepackage{iclr2027_conference}

\usepackage{times}
\usepackage[utf8]{inputenc} 
\usepackage[T1]{fontenc}    
\usepackage[colorlinks=true,urlcolor=black]{hyperref}
\usepackage{url}            
\usepackage{booktabs}       
\usepackage{amsfonts}       
\usepackage{nicefrac}       
\usepackage{microtype}      
\usepackage[table,dvipsnames]{xcolor}
\usepackage{graphicx}
\usepackage{kotex}

\usepackage{amsmath}
\usepackage{amssymb}
\usepackage{mathtools}
\usepackage{amsthm}
\usepackage[capitalize,noabbrev]{cleveref}
\usepackage{enumitem}
\usepackage{algorithm}
\usepackage{algpseudocode}
\usepackage[labelfont=bf]{caption}
\usepackage{subcaption}      
\usepackage{multicol}
\usepackage{multirow}
\usepackage{wrapfig}
\hypersetup{colorlinks,citecolor={teal}}
\usepackage{bbm}
\usepackage{amsmath,amsfonts,bm}

\def\eqref#1{equation~\ref{#1}}

\def\1{\bm{1}}

\DeclareMathAlphabet{\mathsfit}{\encodingdefault}{\sfdefault}{m}{sl}
\SetMathAlphabet{\mathsfit}{bold}{\encodingdefault}{\sfdefault}{bx}{n}

\newcommand{\eg}{\emph{e.g.}}

\newcommand{\D}{\mathcal{D}}

\newcommand{\method}{PRISM}

\theoremstyle{plain}

\theoremstyle{definition}

\theoremstyle{remark}

\algrenewcommand\algorithmicrequire{\textbf{Input:}}
\algrenewcommand\algorithmicensure{\textbf{Output:}}
\algnewcommand\Input{\item[\algorithmicinput]}
\algnewcommand\Output{\it em[\algorithmicoutput]}

\def\hquad{\hskip.5em\relax}

\iclrfinalcopy 
\begin{document}
\maketitle

\begin{abstract}

Irregularly sampled multivariate time series (ISMTS) are prevalent in real-world applications, where both observation times and available measurements can vary substantially across domains.
While recent models increasingly exploit such sampling information for prediction, its robustness under sampling pattern shifts remains underexplored.
We introduce HAR-C, to the best of our knowledge the first controlled benchmark for sampling pattern shifts in ISMTS, and show that sampling shifts alone can substantially degrade performance, induce sampling-specific shortcuts, and remain challenging for existing domain generalization (DG) methods.
Motivated by these findings, we propose \method, a DG framework that first learns complementary feature-centric and sampling-centric representations without task labels, and subsequently performs robust supervised training across diverse sampling variations to discourage brittle shortcut reliance.
Extensive experiments on controlled and real-world ISMTS benchmarks demonstrate that PRISM consistently improves robustness to unseen sampling shifts over existing methods.
Our code is available at \url{https://anonymous.4open.science/r/\method}.
\end{abstract}
\section{Introduction}\label{sec:intro}


Irregularly sampled multivariate time series (ISMTS) are prevalent across various domains, such as electronic health records in hospitals and sensor observations on wearable devices~\citep{azoff1994neural,ismail2019deep}.
Such irregularities stem from various factors, including naturally fluctuating data collection intervals, asynchronous sampling across different features, and uncertainties in observation timestamps~\citep{eckner2012framework,little2019statistical}.
Establishing predictive models in such domains is of significant practical importance yet particularly challenging, since modern deep learning architectures are built upon inductive biases that assume regularly structured inputs (\eg, images, videos, and speech)~\citep{bronstein2017geometric,latentode,mtand}, without explicitly taking into account irregular sampling patterns.

In response to these challenges, earlier approaches largely followed the convention of fitting ISMTS to existing models (\eg, RNNs and Transformers) by forcing irregular observations onto a regular grid through temporal quantization and imputation~\citep{lipton2015learning,grud}.
This incorporates aligning observations to regular intervals and imputing unobserved observations.
However, such forced regularization distorts the original observation times, degrading temporal precision~\citep{latentode}, while imputation introduces estimation errors at unobserved time points.
More importantly, this process discards valuable information about \emph{how observations were acquired}; for instance, deteriorating patients are often monitored more frequently and undergo more tests, making measurement frequency informative of patient condition~\citep{grud}.
These limitations have motivated a shift toward models that directly process irregular time series in their native form, including observation sets (values, feature types, and timestamps)~\citep{seft,mtand,strats}, image-like representations~\citep{vitst}, and token sequences with temporal information~\citep{tan2024language,timellm}.
In essence, modern irregular time series models not only recover the underlying signal, but also preserve and exploit information contained in the observation process.

While this paradigm has substantially advanced ISMTS modeling, its robustness under real-world distribution shifts has yet to be fully scrutinized.
Indeed, sampling patterns frequently vary across domains (\eg, hospitals may differ in vital-sign monitoring frequency or laboratory-test availability), potentially turning useful sampling cues into brittle shortcuts.
To systematically evaluate their impact, we introduce HAR-C, to the best of our knowledge the first controlled benchmark dedicated to sampling pattern shifts in ISMTS, built on human activity recognition (HAR)~\citep{anguita2013public} while preserving the underlying feature distribution and observation budget.
Our analysis reveals that
(i) sampling shifts alone substantially degrade performance, with larger shifts causing greater degradation;
(ii) motivated by the predictive nature of sampling patterns in real-world ISMTS, we synthetically control their correlation with class labels and find that models increasingly follow sampling-associated labels even when they conflict with the underlying signal;
(iii) real-world sampling processes vary substantially across domains while \emph{carrying nontrivial predictive information}, creating conditions for such shortcuts to become unreliable; and
(iv) existing domain generalization (DG) methods~\citep{coral,groupdro,diversify,manydg} remain vulnerable when these shortcuts are stable across source domains but change only at test time.

These observations expose a fundamental dilemma: \emph{sampling information is too useful to discard, yet too domain-dependent to trust unconditionally}; enforcing invariance may remove legitimate predictive cues, whereas unrestricted exploitation encourages brittle dependence on the acquisition process.
To reconcile this tension, we propose \method{}, a model-agnostic DG framework that leverages sampling information while discouraging shortcut reliance.
Since label supervision can directly encourage such shortcuts, \method{} first pre-trains two complementary representations without task labels.
Specifically, two independently parameterized encoders learn a feature-centric representation through feature value reconstruction and a sampling-centric representation through sampling pattern reconstruction.
During labeled training, \method{} constructs multiple stochastic sampling variations for each sample and minimizes the worst classification loss across them, discouraging brittle shortcut reliance while retaining sampling information that remains useful across variations.
Extensive experiments on controlled and real-world ISMTS benchmarks demonstrate improved robustness to unseen sampling shifts over existing DG methods.

In summary, our contributions are threefold:
\begin{itemize}
    \item We introduce HAR-C, a controlled benchmark dedicated to sampling pattern shifts in ISMTS, and provide an in-depth analysis showing that sampling shifts alone degrade performance, induce sampling-specific shortcuts, and remain challenging for existing DG methods.

    \item Motivated by these findings, we propose \method{}, a DG framework that learns complementary feature-centric and sampling-centric representations without task labels and robustly trains across stochastic sampling variations to prevent brittle shortcut reliance.

    \item Extensive experiments on controlled and real-world ISMTS benchmarks demonstrate that \method{} consistently improves robustness to unseen sampling shifts over existing DG methods, supported by comprehensive ablation and representation analyses.
\end{itemize}
\section{In-depth Analysis of Sampling Pattern Shifts}\label{sec:analysis}

\begin{wrapfigure}{r}{0.4\textwidth}
\centering
\vspace{-.1in}
\includegraphics[width=\linewidth]{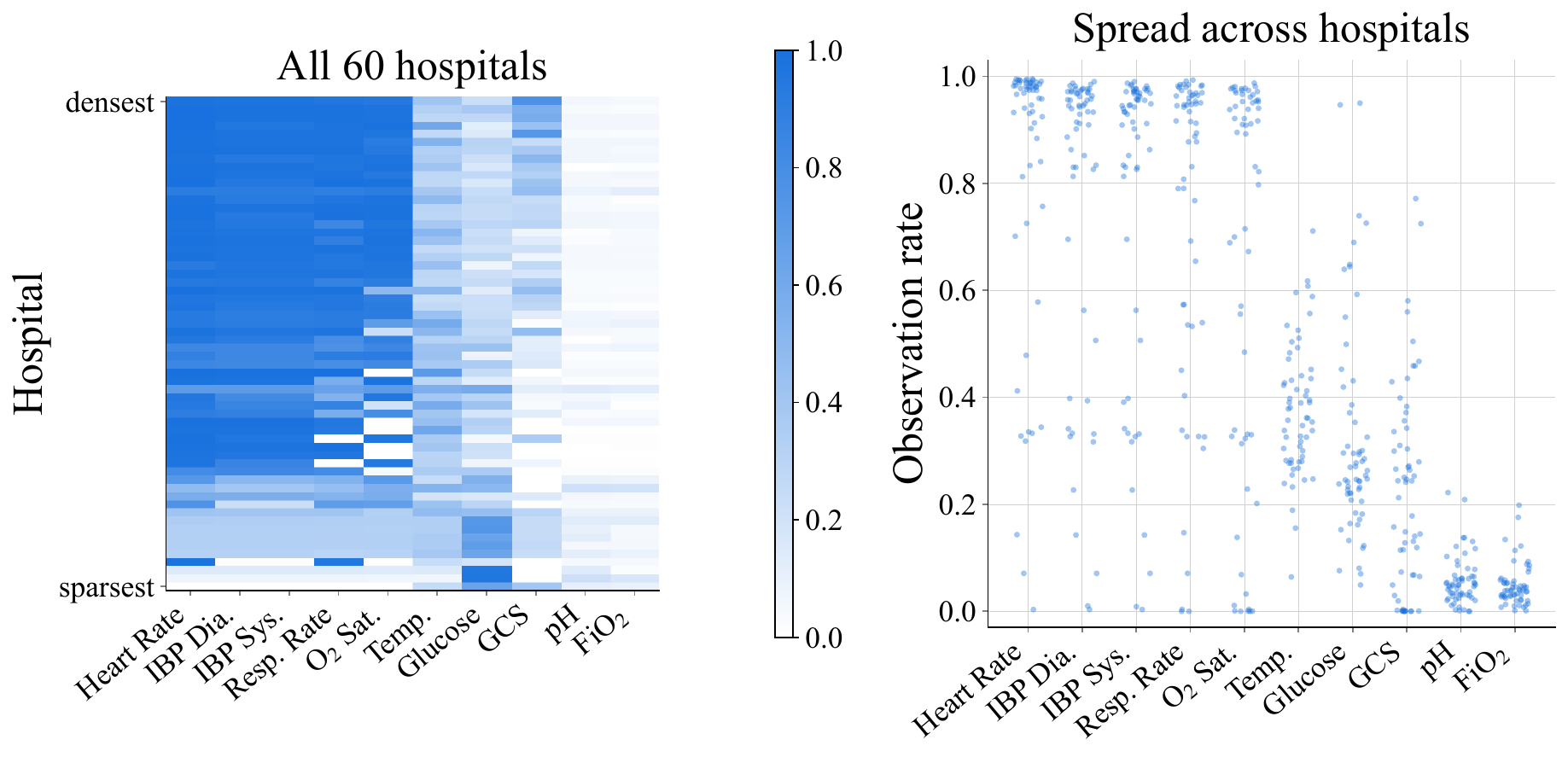}
\vspace{-.25in}
\caption{
Hospital-level sampling pattern variations on eICU dataset. 
}
\label{fig:eicu_hospital_masks}
\vspace{-.2in}
\end{wrapfigure}

In this section, we analyze how sampling pattern shifts affect prediction, induce shortcut reliance, arise in real-world data, and challenge existing domain generalization methods.
Real-world time series often exhibit irregularities arising from opportunistic data collection, sensor failures, or event-driven observations, as illustrated in~\Cref{fig:eicu_hospital_masks}, where feature observation rates in eICU~\citep{eicu} vary substantially across hospitals.
Importantly, the sampling process itself may vary across domains, changing \emph{how} the same underlying signal is observed.

\subsection{HAR-C: A Controlled Benchmark for Sampling Pattern Shifts}
\label{sec:analysis_harc}

Studying the isolated effect of sampling pattern shifts is difficult since changing the observation process often also changes the amount of available information or introduces broader distribution shifts.
To disentangle these factors, we introduce \textit{HAR-C}, a controlled benchmark built on the standard Human Activity Recognition (HAR) dataset~\citep{anguita2013public}, where sampling patterns are manipulated while preserving the underlying signals and observation budget.

\begin{figure*}[!t]
\centering
\includegraphics[width=\textwidth]{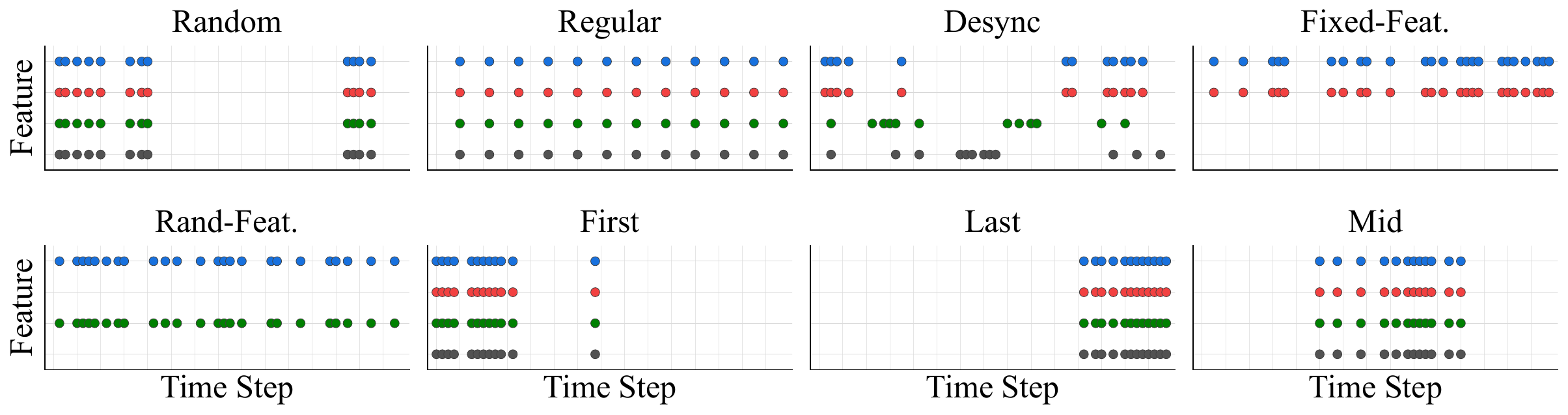}
\vspace{-.2in}
\caption{
Overview of sampling pattern shifts in HAR-C.
Starting from synchronized random sampling, HAR-C constructs diverse target sampling patterns while preserving the observation budget.
}
\label{fig:harc_construction}
\vspace{-.2in}
\end{figure*}

Starting from regularly sampled HAR sequences, we retain $20\%$ of the original observations to construct irregular inputs.
Since the sensor channels in HAR are jointly recorded at each time point, the source domain, Random, uses synchronized random sampling across features.
As illustrated in~\Cref{fig:harc_construction}, we construct diverse target shifts, including regular sampling (Regular, reference), desynchronization across features (Desync), heterogeneous feature-wise sampling (Fixed-Feat. and Rand-Feat.), and temporal concentration toward different parts of the sequence (First, Last, and Mid), while varying shift severity.
Crucially, all sampling patterns preserve the same 20\% observation budget, allowing us to separate the effect of sampling pattern shifts from that of a reduced observation budget.

\subsection{Analysis of Sampling Pattern Shifts}
\label{sec:analysis_sampling}

\begin{figure*}[!t]
\centering
\includegraphics[width=\textwidth]{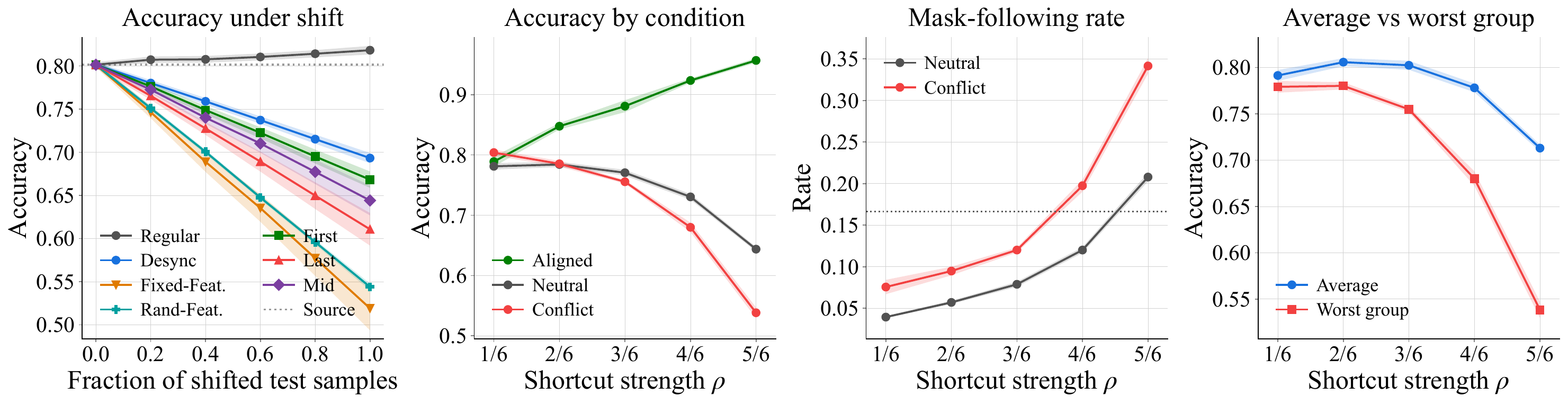}
\vspace{-.25in}
\caption{
Analysis of sampling pattern shifts on HAR-C. The four panels show accuracy under increasing shift severity, accuracy under varying shortcut strength $\rho$, mask-following rate, and average versus worst-group accuracy.
}
\label{fig:analysis_severity_shortcut}
\vspace{-.2in}
\end{figure*}

\paragraph{Sampling pattern shifts alone substantially degrade prediction.}
We first examine how performance changes as an increasing fraction of test samples undergo each sampling shift.
Throughout this analysis, we use a CNN with zero imputation as a simple and commonly used baseline for irregularly observed time series~\citep{grud,kim2023probabilistic,lee2026deep}.
As shown in \Cref{fig:analysis_severity_shortcut}, performance deteriorates under target sampling shifts despite preserving the same observation budget, with larger fractions of shifted samples generally causing greater degradation.
The Regular process serves as a reference condition, whereas the remaining shifts reveal substantial sensitivity to \emph{where} and \emph{how} observations are collected.
This behavior also persists across SeFT~\citep{seft}, mTAND~\citep{mtand}, and GRU-D~\citep{grud} in~\Cref{fig:a1_backbones}, indicating that the phenomenon is not architecture-specific.

\paragraph{Sampling patterns become shortcuts when correlated with labels.}
We next investigate whether models exploit sampling patterns as predictive shortcuts.
We associate each of the six HAR classes with a distinct sampling pattern and vary the sampling--label association strength $\rho$ from chance level ($1/6$) to increasingly predictive settings.
At test time, the sampling cue is aligned with the true label, neutral, or conflicting.
As shown in \Cref{fig:analysis_severity_shortcut}, increasing $\rho$ improves aligned accuracy but sharply degrades conflicting accuracy, while the rising mask-following rate shows that predictions increasingly track the sampling-associated label.
The widening average--worst-group gap further reveals failures hidden by strong average performance.
A likely reason is that sampling patterns provide simple, directly observable cues, whereas class-relevant feature information requires integrating multiple noisy observations and their temporal dynamics.

\begin{figure*}[!t]
\centering
\includegraphics[width=\textwidth]{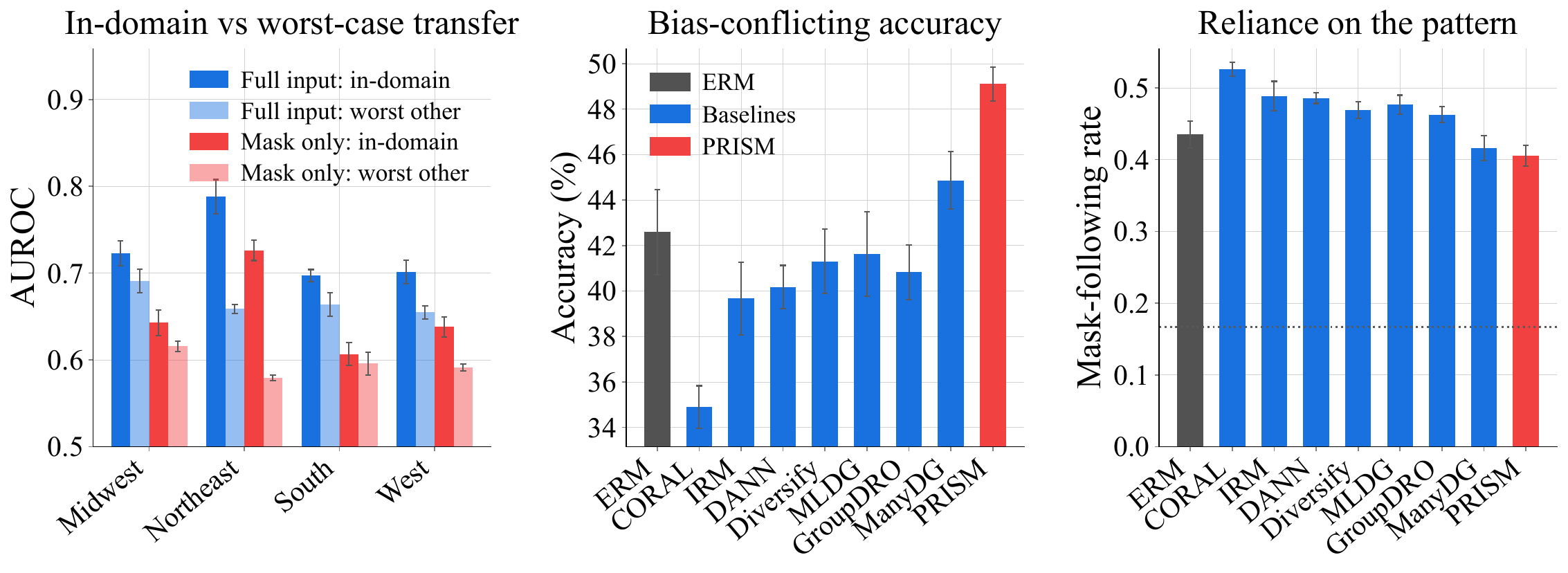}
\vspace{-.3in}
\caption{
Analysis of sampling-specific shortcuts and domain generalization. The three panels show in-domain versus worst cross-region transfer on eICU, bias-conflicting accuracy on HAR-C, and mask-following rate across DG methods.
}
\label{fig:dg_realworld_shortcut}
\vspace{-.25in}
\end{figure*}

\paragraph{Real-world sampling processes are predictive, yet domain dependent.}
We next examine whether the same phenomenon arises in inherently irregular real-world electronic health records, using mortality prediction on eICU~\citep{eicu}.
Since deteriorating patients are often monitored more frequently or receive additional measurements, the observation process can correlate with patient condition.
Indeed, the leftmost panel of \Cref{fig:dg_realworld_shortcut} shows that observation masks alone achieve AUROC substantially above $0.5$, indicating that \emph{sampling patterns themselves contain predictive information} about mortality.
However, this relationship is \emph{unstable across environments}: observation frequencies vary substantially across hospitals (\Cref{fig:eicu_hospital_masks}), and mask-only prediction degrades markedly under cross-region transfer.
Thus, sampling patterns can be both predictive and domain dependent---the two ingredients that make sampling-specific shortcuts brittle under domain shift.

\paragraph{Existing DG remains vulnerable to source-stable sampling shortcuts.}
Finally, we ask whether existing domain generalization methods can address this failure mode.
We construct source environments in which the sampling--label association remains stable and predictive, but reverses in the unseen target domain.
We evaluate bias-conflicting accuracy and the mask-following rate, measuring whether predictions follow the true class or the misleading sampling cue when the two conflict.
As shown in \Cref{fig:dg_realworld_shortcut}, existing DG methods provide only limited improvements over ERM while continuing to rely substantially on the sampling pattern.
Under a strong sampling--label association, they perform well when the shortcut agrees with the label but degrade sharply when it becomes conflicting, suggesting that sampling patterns can remain a dominant shortcut under standard DG training.

\section{Proposed method: \method{}}\label{sec:method}

\begin{figure*}[!t]
\centering
\includegraphics[width=\textwidth]{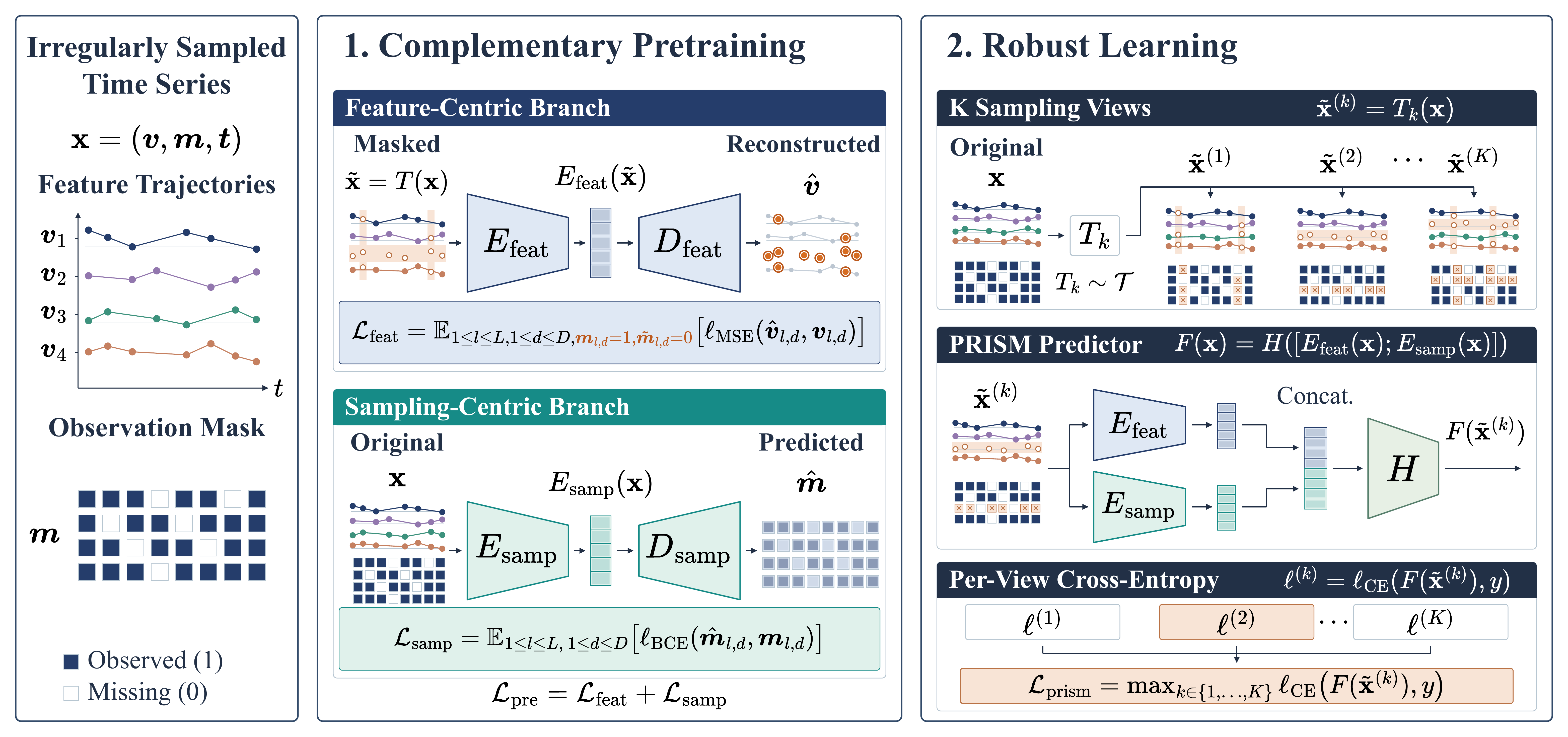}
\vspace{-.25in}
\caption{
Overview of \method{}.
Feature- and sampling-centric encoders are first pretrained through masked value and observation-mask reconstruction, then jointly trained using the worst classification loss across stochastic sampling variations.
}
\label{fig:concept}
\vspace{-.2in}
\end{figure*}

This section elucidates \method{}, a model-agnostic DG framework for learning robust representations under sampling pattern shifts in ISMTS.
Motivated by analysis in~\Cref{sec:analysis}, \method{} first pretrains complementary feature-centric and sampling-centric representations through value reconstruction and sampling pattern reconstruction, respectively, without task labels (\Cref{sec:method_pretrain}).
Second, the two representations are jointly used for downstream prediction, where robust optimization across stochastically generated sampling variations discourages brittle reliance on sampling-specific shortcuts (\Cref{sec:method_robust}).
An overview of \method{} is illustrated in~\Cref{fig:concept}.

\subsection{Problem Setup}\label{sec:method_setup}

Before delving into the details of \method{}, we first formally define the problem setup of DG on ISMTS.
Let $\mathcal{X}$ be the space of valid ISMTS inputs and $\mathcal{Y} = \{1, \dots, C\}$ the label space for $C$-class classification.
Each $\mathbf{x} \in \mathcal{X}$ is represented as $\mathbf{x} = (\boldsymbol{v}, \boldsymbol{m}, \boldsymbol{t})$, where, for a sequence with $L$ time points and $D$ features, $\boldsymbol{v} \in \mathbb{R}^{L \times D}$ denotes feature values, $\boldsymbol{m} \in \{0, 1\}^{L \times D}$ is the corresponding observation mask (0: missing, 1: observed), and $\boldsymbol{t} \in \mathbb{R}^{L \times 1}$ denotes observation times.
Given labeled source data $\D_s \subseteq \mathcal{X} \times \mathcal{Y}$, our goal is to learn a predictor $F: \mathcal{X} \rightarrow \Delta^{C-1}$ that generalizes to an unseen target domain $\D_t \subseteq \mathcal{X} \times \mathcal{Y}$, whose sampling process may differ from those observed during training, where $\Delta^{C-1}$ denotes the probability simplex over $C$ classes.

\subsection{Complementary Representation Pretraining}\label{sec:method_pretrain}

Our analysis in~\Cref{sec:analysis} reveals that sampling patterns are informative yet prone to shortcut reliance under task supervision.
This creates a tension: learning feature representations invariant to sampling patterns may discard useful predictive cues, whereas learning a single representation from both feature values and sampling patterns may still rely on sampling-specific shortcuts.
\method{} resolves this tension by pretraining two complementary representations without task labels, separately focusing on feature values and sampling patterns.

Given the backbone architecture of $F$, we instantiate two independently parameterized encoders: a feature-centric encoder $E_{\mathrm{feat}}$ and a sampling-centric encoder $E_{\mathrm{samp}}$.
This model-agnostic design applies across different ISMTS architectures.
First, the feature-centric encoder is encouraged to capture rich information about the observed feature values.
We sample a transformation $T \sim \mathcal{T}$ and construct a partially observed view $\tilde{\mathbf{x}} = T(\mathbf{x})$ by randomly hiding additional observations.
From the remaining measurements, $E_{\mathrm{feat}}$ must retain sufficient information to recover the masked values.
A lightweight decoder $D_{\mathrm{feat}}$ produces $\hat{\boldsymbol{v}} = D_{\mathrm{feat}}(E_{\mathrm{feat}}(\tilde{\mathbf{x}})) \in \mathbb{R}^{L \times D}$.
The reconstruction loss is evaluated only on observations originally present but removed by $T$:
\begin{equation}\label{eq:feature_recon}
    \mathcal{L}_{\mathrm{feat}}
    =
    \mathbb{E}_{1 \le l \le L, \, 1 \le d \le D, \,
    \boldsymbol{m}_{l, d}=1, \,
    \tilde{\boldsymbol{m}}_{l, d}=0}
    \left[
        \ell_{\mathrm{MSE}}
        \left(
            \hat{\boldsymbol{v}}_{l, d},
            \boldsymbol{v}_{l, d}
        \right)
    \right].
\end{equation}
This masked recovery task prevents simple input copying and encourages the representation to capture informative structure across the observed measurements.

The sampling-centric encoder instead preserves how the time series was observed.
Since sampling patterns can themselves carry predictive information, a lightweight decoder $D_{\mathrm{samp}}$ reconstructs the original observation mask from $E_{\mathrm{samp}}(\mathbf{x})$, yielding $\hat{\boldsymbol{m}} = D_{\mathrm{samp}}(E_{\mathrm{samp}}(\mathbf{x})) \in (0, 1)^{L \times D}$.
Although $E_{\mathrm{samp}}$ receives the full input, reconstructing m through a compact representation encourages it to summarize the observation structure rather than copy the input. Its reconstruction objective is
\begin{equation}\label{eq:sampling_recon}
    \mathcal{L}_{\mathrm{samp}}
    =
    \mathbb{E}_{1 \le l \le L, \, 1 \le d \le D}
    \left[
        \ell_{\mathrm{BCE}}
        \left(
            \hat{\boldsymbol{m}}_{l, d},
            \boldsymbol{m}_{l, d}
        \right)
    \right].
\end{equation}
Unlike $\mathcal{L}_{\mathrm{feat}}$, $\mathcal{L}_{\mathrm{samp}}$ requires no additional masking, as its goal is to preserve rather than infer the sampling structure.
Here, $\ell_{\mathrm{MSE}}(a,b)=(a-b)^2$ and $\ell_{\mathrm{BCE}}$ denote the mean-squared error and binary cross-entropy loss, respectively.
Together, the two branches retain complementary views of the input---what is observed and how it is observed---without task labels shaping either representation.
The two reconstruction objectives are equally weighted,
$\mathcal{L}_{\mathrm{pre}} = \mathcal{L}_{\mathrm{feat}} + \mathcal{L}_{\mathrm{samp}}$,
avoiding an additional balancing hyperparameter.
After pretraining, the lightweight decoders $D_{\mathrm{feat}}$ and $D_{\mathrm{samp}}$ are discarded, leaving the two pretrained encoders for downstream prediction~\citep{he2022masked}.

The augmentation distribution $\mathcal{T}$ is intentionally simple; it is not designed to mimic any particular target shift, and its parameters are selected using source-domain validation data only.
A transformation $T \sim \mathcal{T}$ independently samples timestep and feature drop ratios
$p_t \sim \mathcal{U}(0,r_t)$ and $p_f \sim \mathcal{U}(0,r_f)$, and randomly masks the selected time points and features.
The same augmentation distribution is reused to create diverse sampling variations for downstream robust learning.

\subsection{Robust Learning Across Sampling Variations}
\label{sec:method_robust}

The complementary representations learned during pretraining do not by themselves prevent shortcut reliance during downstream supervision.
If simply concatenated, the classifier may still favor sampling cues that are highly predictive in the source data, while fixing their relative contributions is undesirable because their importance can vary across samples.
We therefore combine both representations flexibly while discouraging dependencies that are fragile to sampling pattern changes.

For an input $\mathbf{x}$, the two representations are concatenated and passed through a shared classification head $H$:
$F(\mathbf{x}) = H([E_{\mathrm{feat}}(\mathbf{x}); E_{\mathrm{samp}}(\mathbf{x})])$.
Since the sampling process of the unseen target domain $\D_t$ is unknown, we expose each training sample to multiple plausible sampling variations.
Specifically, we draw $K$ transformations $T_k \sim \mathcal{T}$ and construct $\tilde{\mathbf{x}}^{(k)} = T_k(\mathbf{x})$ for $k \in \{1, \dots, K\}$, each preserving the underlying instance and label while altering its sampling pattern.

For each sample, \method{} defines its robust classification loss as the worst loss among $K$ sampling variations:
\begin{equation}
\label{eq:robust_objective}
    \mathcal{L}_{\mathrm{prism}}
    =
    \max_{k \in \{1, \dots, K\}}
    \ell_{\mathrm{CE}}
    \left(
        F(\tilde{\mathbf{x}}^{(k)}), y
    \right),
\end{equation}
where $\ell_{\mathrm{CE}}$ denotes the cross-entropy loss.
The objective is averaged over source samples, yielding sample-wise minimax optimization over sampling variations.

If the predictor relies on a sampling-specific shortcut, some augmented views can disrupt that cue and incur high loss.
The worst-view objective thus prevents such failures from being averaged out, allowing the predictor to flexibly exploit feature and sampling information without an additional balancing hyperparameter or gating network.

\section{Experiments}

In this section, we thoroughly evaluate \method{} through the following research questions:
(\textbf{RQ1}) Does \method{} robustly generalize to unseen sampling pattern shifts across controlled, real-world, and increasingly severe settings?~(\Cref{exp:main_result});
(\textbf{RQ2}) Which design components are responsible for this robustness?~(\Cref{exp:ablation_study});
(\textbf{RQ3}) Does the robustness generalize across backbone architectures while remaining practical?~(\Cref{exp:further_analysis})

\subsection{Experimental setup}\label{exp:setup}

\paragraph{Datasets.}
We evaluate \method{} on three complementary benchmarks covering controlled, naturally occurring, and increasingly severe sampling shifts.
HAR-C, introduced in~\Cref{sec:analysis_harc}, isolates sampling pattern shifts under a fixed observation budget, allowing us to evaluate their effect without confounding changes in the amount of observed information.
eICU~\citep{eicu} considers in-hospital mortality prediction under leave-one-region-out shifts across four regions, where sampling pattern shifts are intertwined with broader naturally occurring distribution shifts.
PAM~\citep{pam} contains multivariate sensor data from multiple individuals and follows~\citep{raindrop,vitst} with leave-fixed-sensors-out shifts of increasing severity.
Together, these benchmarks test whether robustness observed under controlled shifts transfers to more realistic and severe changes in the observation process.
Detailed dataset statistics are provided in~\Cref{tab:dataset-statistics}.

\paragraph{Domain generalization baselines.}
We compare \method{} with ERM and representative DG baselines: ARM~\citep{arm}, CORAL~\citep{coral}, DANN~\citep{dann}, GroupDRO~\citep{groupdro}, IB-ERM~\citep{ib_erm}, IRM~\citep{irm}, Mixup~\citep{mixup}, MLDG~\citep{mldg}, MMD~\citep{mmd}, VREx~\citep{vrex}, Diversify~\citep{diversify}, and ManyDG~\citep{manydg}.
These baselines span widely used general-purpose DG methods as well as approaches specifically designed for time series, covering diverse strategies including data augmentation, invariant representation learning, robust optimization, and meta-learning.

\paragraph{Implementation details.}
Following prior time series DG work~\citep{diversify}, we primarily use a CNN backbone, with alternative backbones evaluated in~\Cref{sec:additional_experiments}.
We use $K=4$ throughout, with dataset-specific sampling augmentations detailed in~\Cref{tab:augmentation_settings}.
All experiments are repeated ten times, and we report the mean$\pm$standard error. 
Further implementation details and configurations are provided in the appendix and at \url{https://anonymous.4open.science/r/PRISM}.

\begin{table*}[!t]
\centering
\caption{
Comparison of \method{} and existing DG methods on HAR-C.
Models are evaluated on the source-matched Random condition and seven target sampling pattern shifts.
}
\label{tab:harc}
\vspace{-.1in}
\resizebox{\textwidth}{!}{
\setlength{\tabcolsep}{4pt}
\begin{tabular}{l|cccccccc|c}
\toprule

\textbf{Algorithm} & \textbf{Random} & \textbf{Regular} & \textbf{Desync} & \textbf{Fixed-Feat.} & \textbf{Rand-Feat.} & \textbf{First} & \textbf{Last} & \textbf{Mid} & \textbf{Avg.} \\

\midrule

ERM & 79.92\scriptsize{$\pm$0.55} & 80.97\scriptsize{$\pm$0.81} & 68.91\scriptsize{$\pm$0.98} & 54.05\scriptsize{$\pm$4.27} & 54.48\scriptsize{$\pm$0.26} & 67.25\scriptsize{$\pm$1.52} & 61.10\scriptsize{$\pm$3.20} & 65.06\scriptsize{$\pm$2.50} & 66.47\scriptsize{$\pm$0.95} \\

ARM & 80.39\scriptsize{$\pm$0.49} & 81.32\scriptsize{$\pm$0.82} & 68.15\scriptsize{$\pm$0.92} & 56.54\scriptsize{$\pm$3.56} & 56.74\scriptsize{$\pm$0.65} & 73.12\scriptsize{$\pm$0.88} & 70.54\scriptsize{$\pm$0.87} & 72.18\scriptsize{$\pm$1.04} & 69.87\scriptsize{$\pm$0.60} \\

CORAL & 80.05\scriptsize{$\pm$0.54} & 81.51\scriptsize{$\pm$0.67} & 70.26\scriptsize{$\pm$1.22} & 53.97\scriptsize{$\pm$3.49} & 54.91\scriptsize{$\pm$0.48} & 67.12\scriptsize{$\pm$1.90} & 60.34\scriptsize{$\pm$3.60} & 67.10\scriptsize{$\pm$2.86} & 66.91\scriptsize{$\pm$0.88} \\

DANN & 81.94\scriptsize{$\pm$0.24} & 82.72\scriptsize{$\pm$0.73} & 70.30\scriptsize{$\pm$0.90} & 51.82\scriptsize{$\pm$4.97} & 54.41\scriptsize{$\pm$0.53} & 69.21\scriptsize{$\pm$1.25} & 60.59\scriptsize{$\pm$2.66} & 68.12\scriptsize{$\pm$1.20} & 67.39\scriptsize{$\pm$0.98} \\

GroupDRO & 79.86\scriptsize{$\pm$0.21} & 81.41\scriptsize{$\pm$0.55} & 68.45\scriptsize{$\pm$0.84} & 54.79\scriptsize{$\pm$3.99} & 55.01\scriptsize{$\pm$0.43} & 66.53\scriptsize{$\pm$1.08} & 61.43\scriptsize{$\pm$2.15} & 66.78\scriptsize{$\pm$1.51} & 66.78\scriptsize{$\pm$0.68} \\

IB\_ERM & 80.51\scriptsize{$\pm$0.45} & 81.45\scriptsize{$\pm$0.84} & 68.94\scriptsize{$\pm$0.97} & 53.87\scriptsize{$\pm$4.17} & 56.10\scriptsize{$\pm$0.53} & 64.53\scriptsize{$\pm$1.34} & 61.17\scriptsize{$\pm$3.46} & 66.10\scriptsize{$\pm$2.81} & 66.58\scriptsize{$\pm$1.26} \\

IRM & 80.53\scriptsize{$\pm$0.41} & 81.55\scriptsize{$\pm$0.47} & 68.27\scriptsize{$\pm$0.64} & 53.52\scriptsize{$\pm$4.62} & 53.74\scriptsize{$\pm$0.65} & 67.35\scriptsize{$\pm$1.66} & 61.16\scriptsize{$\pm$2.31} & 66.02\scriptsize{$\pm$1.81} & 66.52\scriptsize{$\pm$1.11} \\

Mixup & 80.01\scriptsize{$\pm$1.08} & 81.30\scriptsize{$\pm$1.75} & 65.46\scriptsize{$\pm$2.47} & 55.45\scriptsize{$\pm$4.78} & 54.47\scriptsize{$\pm$0.61} & 68.93\scriptsize{$\pm$1.96} & 64.06\scriptsize{$\pm$2.18} & 71.12\scriptsize{$\pm$1.19} & 67.60\scriptsize{$\pm$1.17} \\

MLDG & 84.01\scriptsize{$\pm$0.29} & 86.08\scriptsize{$\pm$0.18} & 75.38\scriptsize{$\pm$0.68} & 58.85\scriptsize{$\pm$3.98} & 56.54\scriptsize{$\pm$0.44} & 74.66\scriptsize{$\pm$0.87} & 71.22\scriptsize{$\pm$0.72} & 74.84\scriptsize{$\pm$0.74} & 72.70\scriptsize{$\pm$0.54} \\

MMD & 80.96\scriptsize{$\pm$0.24} & 83.02\scriptsize{$\pm$0.68} & 70.34\scriptsize{$\pm$0.94} & 54.36\scriptsize{$\pm$4.51} & 54.32\scriptsize{$\pm$0.63} & 69.23\scriptsize{$\pm$1.15} & 67.69\scriptsize{$\pm$1.80} & 71.10\scriptsize{$\pm$1.59} & 68.88\scriptsize{$\pm$0.85} \\

VREx & 70.41\scriptsize{$\pm$1.08} & 71.05\scriptsize{$\pm$1.66} & 58.42\scriptsize{$\pm$1.55} & 50.30\scriptsize{$\pm$4.53} & 52.27\scriptsize{$\pm$0.55} & 63.50\scriptsize{$\pm$1.55} & 64.68\scriptsize{$\pm$1.10} & 65.20\scriptsize{$\pm$1.41} & 61.98\scriptsize{$\pm$1.05} \\

Diversify & 82.76\scriptsize{$\pm$0.31} & 84.19\scriptsize{$\pm$0.58} & 71.04\scriptsize{$\pm$1.14} & 53.71\scriptsize{$\pm$4.64} & 53.86\scriptsize{$\pm$0.46} & 70.88\scriptsize{$\pm$0.91} & 65.31\scriptsize{$\pm$1.28} & 71.70\scriptsize{$\pm$0.65} & 69.18\scriptsize{$\pm$0.75} \\

ManyDG & 82.18\scriptsize{$\pm$0.49} & 84.85\scriptsize{$\pm$0.63} & 73.06\scriptsize{$\pm$0.88} & 51.66\scriptsize{$\pm$5.33} & 55.92\scriptsize{$\pm$0.80} & 66.44\scriptsize{$\pm$2.45} & 61.87\scriptsize{$\pm$1.80} & 68.64\scriptsize{$\pm$1.96} & 68.08\scriptsize{$\pm$1.16} \\

\midrule

\rowcolor{gray!20} \method{} & \textbf{84.08\scriptsize{$\pm$0.16}} & \textbf{86.49\scriptsize{$\pm$0.21}} & \textbf{82.81\scriptsize{$\pm$0.27}} & \textbf{74.90\scriptsize{$\pm$1.29}} & \textbf{74.72\scriptsize{$\pm$0.21}} & \textbf{81.28\scriptsize{$\pm$0.20}} & \textbf{79.22\scriptsize{$\pm$0.35}} & \textbf{81.02\scriptsize{$\pm$0.23}} & \textbf{80.57\scriptsize{$\pm$0.19}} \\

\bottomrule
\end{tabular}}
\vspace{-.2in}
\end{table*}

\subsection{Main results}\label{exp:main_result}

\paragraph{Controlled sampling pattern shifts.}
To answer RQ1, we first evaluate \method{} on HAR-C, where only the sampling process changes while the observation budget is controlled.
\method{} achieves the best performance across all eight sampling conditions, improving average accuracy from $72.70\%$ for the strongest baseline to $80.57\%$~(\Cref{tab:harc}).
Excluding the Random and Regular conditions, the gap widens further (78.99\% vs. 68.58\%).
Importantly, under the source-matched Random condition, \method{} and MLDG perform nearly identically ($84.08\%$ vs.\ $84.01\%$), whereas substantially larger gaps emerge once the target sampling process differs from that seen during training.
Relative to the strongest baseline, this suggests that the gains stem mainly from maintaining performance as the observation process changes, rather than from better fitting the source distribution.
Consistent with this interpretation, \method{} also achieves the strongest unbiased and bias-conflicting accuracy when sampling-specific shortcuts are introduced (\Cref{tab:harc-bias}), while its mask-following rate remains comparable to ERM (\Cref{fig:dg_realworld_shortcut}), suggesting that it makes better use of feature information rather than discarding sampling cues.

Overall, the controlled setting directly attributes the gains of \method{} to robustness against sampling pattern shifts.

\paragraph{Natural and increasingly severe sampling shifts.}
We next test whether this robustness extends beyond the controlled HAR-C setting.
On eICU, \method{} achieves the strongest AUROC and AUPRC across all four target regions despite heterogeneous real-world shifts, including changes in sampling patterns~(\Cref{tab:eicu}).
In contrast, existing DG baselines provide only limited gains over ERM, suggesting that standard DG objectives do not sufficiently address variation in the observation process when it co-occurs with broader distribution shifts.
The consistent gains of \method{} across regions therefore indicate that explicitly accounting for sampling variability remains beneficial even when sampling shifts cannot be isolated from other sources of domain variation.

On PAM, the difference becomes increasingly pronounced as the target sampling process departs further from the source.
While performance remains relatively stable under mild sensor removal, \method{} maintains $81.97\%$ accuracy when $50\%$ of sensors are entirely unavailable, compared with $58.99\%$ for the strongest baseline~(\Cref{tab:pam_lso}).
The growing gap with increasing removal severity suggests that the benefit of \method{} becomes larger precisely when the available observation structure differs most strongly from training.
Together, eICU and PAM show that the robustness observed on HAR-C transfers to naturally occurring shifts and persists under substantially more severe changes in sampling.

\begin{table*}[!t]
\centering
\caption{
Comparison of \method{} and existing DG methods on eICU under leave-one-region-out evaluation.
AUROC and AUPRC are reported for Northeast, West, South, and Midwest target regions.
}
\label{tab:eicu}
\vspace{-.1in}
\resizebox{\textwidth}{!}{
\setlength{\tabcolsep}{4pt}
\begin{tabular}{l|cc|cc|cc|cc|cc}
\toprule

\multirow{2}{*}[-0.2em]{\textbf{Algorithm}} & \multicolumn{2}{c}{\textbf{Northeast}} & \multicolumn{2}{c}{\textbf{West}} & \multicolumn{2}{c}{\textbf{South}} & \multicolumn{2}{c}{\textbf{Midwest}} & \multicolumn{2}{c}{\textbf{Avg.}} \\
\cmidrule(l{3pt}r{3pt}){2-3} \cmidrule(l{3pt}r{3pt}){4-5} \cmidrule(l{3pt}r{3pt}){6-7} \cmidrule(l{3pt}r{3pt}){8-9} \cmidrule(l{3pt}r{3pt}){10-11}
 & AUROC & AUPRC & AUROC & AUPRC & AUROC & AUPRC & AUROC & AUPRC & AUROC & AUPRC \\

\midrule

ERM & 85.30\scriptsize{$\pm$0.20} & 57.91\scriptsize{$\pm$0.35} & 76.89\scriptsize{$\pm$0.20} & 43.77\scriptsize{$\pm$0.19} & 78.59\scriptsize{$\pm$0.20} & 40.96\scriptsize{$\pm$0.28} & 78.09\scriptsize{$\pm$0.34} & 37.27\scriptsize{$\pm$0.98} & 79.72\scriptsize{$\pm$0.15} & 44.98\scriptsize{$\pm$0.25} \\

ARM & 74.08\scriptsize{$\pm$1.25} & 45.82\scriptsize{$\pm$0.99} & 68.76\scriptsize{$\pm$1.18} & 37.86\scriptsize{$\pm$1.05} & 69.96\scriptsize{$\pm$1.30} & 33.90\scriptsize{$\pm$0.76} & 72.64\scriptsize{$\pm$0.48} & 31.48\scriptsize{$\pm$0.47} & 71.36\scriptsize{$\pm$0.60} & 37.26\scriptsize{$\pm$0.43} \\

CORAL & 85.08\scriptsize{$\pm$0.24} & 58.16\scriptsize{$\pm$0.27} & 76.99\scriptsize{$\pm$0.29} & 43.75\scriptsize{$\pm$0.33} & 79.47\scriptsize{$\pm$0.18} & 41.97\scriptsize{$\pm$0.20} & 78.58\scriptsize{$\pm$0.14} & 39.25\scriptsize{$\pm$0.25} & 80.03\scriptsize{$\pm$0.08} & 45.78\scriptsize{$\pm$0.11} \\

DANN & 83.28\scriptsize{$\pm$0.31} & 55.97\scriptsize{$\pm$0.40} & 77.90\scriptsize{$\pm$0.18} & 44.64\scriptsize{$\pm$0.35} & 77.38\scriptsize{$\pm$0.29} & 40.04\scriptsize{$\pm$0.36} & 77.19\scriptsize{$\pm$0.31} & 35.29\scriptsize{$\pm$0.57} & 78.94\scriptsize{$\pm$0.16} & 43.98\scriptsize{$\pm$0.24} \\

GroupDRO & 84.36\scriptsize{$\pm$0.30} & 57.40\scriptsize{$\pm$0.27} & 76.99\scriptsize{$\pm$0.27} & 43.85\scriptsize{$\pm$0.33} & 77.33\scriptsize{$\pm$0.33} & 39.32\scriptsize{$\pm$0.36} & 77.39\scriptsize{$\pm$0.39} & 37.76\scriptsize{$\pm$0.45} & 79.02\scriptsize{$\pm$0.13} & 44.58\scriptsize{$\pm$0.16} \\

IB\_ERM & 85.38\scriptsize{$\pm$0.22} & 57.81\scriptsize{$\pm$0.29} & 76.91\scriptsize{$\pm$0.45} & 43.69\scriptsize{$\pm$0.54} & 78.41\scriptsize{$\pm$0.30} & 40.79\scriptsize{$\pm$0.42} & 77.69\scriptsize{$\pm$0.39} & 36.49\scriptsize{$\pm$0.96} & 79.60\scriptsize{$\pm$0.11} & 44.70\scriptsize{$\pm$0.23} \\

IRM & 85.11\scriptsize{$\pm$0.21} & 57.62\scriptsize{$\pm$0.38} & 75.92\scriptsize{$\pm$0.65} & 42.84\scriptsize{$\pm$0.70} & 78.42\scriptsize{$\pm$0.21} & 40.99\scriptsize{$\pm$0.35} & 77.42\scriptsize{$\pm$0.41} & 37.17\scriptsize{$\pm$0.60} & 79.22\scriptsize{$\pm$0.18} & 44.65\scriptsize{$\pm$0.21} \\

Mixup & 85.01\scriptsize{$\pm$0.26} & 57.80\scriptsize{$\pm$0.19} & 78.32\scriptsize{$\pm$0.26} & 45.31\scriptsize{$\pm$0.34} & 79.06\scriptsize{$\pm$0.25} & 41.65\scriptsize{$\pm$0.33} & 78.61\scriptsize{$\pm$0.19} & 38.24\scriptsize{$\pm$0.53} & 80.25\scriptsize{$\pm$0.11} & 45.75\scriptsize{$\pm$0.18} \\

MLDG & 85.81\scriptsize{$\pm$0.21} & 58.95\scriptsize{$\pm$0.31} & 77.57\scriptsize{$\pm$0.30} & 44.49\scriptsize{$\pm$0.49} & 78.87\scriptsize{$\pm$0.11} & 41.08\scriptsize{$\pm$0.30} & 78.31\scriptsize{$\pm$0.24} & 38.25\scriptsize{$\pm$0.41} & 80.14\scriptsize{$\pm$0.07} & 45.70\scriptsize{$\pm$0.14} \\

MMD & 83.99\scriptsize{$\pm$0.58} & 55.97\scriptsize{$\pm$0.57} & 75.71\scriptsize{$\pm$0.59} & 43.02\scriptsize{$\pm$0.71} & 72.66\scriptsize{$\pm$1.29} & 37.19\scriptsize{$\pm$0.84} & 77.08\scriptsize{$\pm$0.41} & 37.57\scriptsize{$\pm$0.52} & 77.36\scriptsize{$\pm$0.46} & 43.44\scriptsize{$\pm$0.40} \\

VREx & 81.52\scriptsize{$\pm$1.01} & 52.80\scriptsize{$\pm$1.04} & 73.06\scriptsize{$\pm$0.80} & 39.04\scriptsize{$\pm$1.32} & 76.26\scriptsize{$\pm$0.28} & 38.03\scriptsize{$\pm$0.40} & 75.73\scriptsize{$\pm$0.22} & 34.62\scriptsize{$\pm$0.54} & 76.64\scriptsize{$\pm$0.32} & 41.12\scriptsize{$\pm$0.45} \\

Diversify & 84.11\scriptsize{$\pm$0.28} & 56.09\scriptsize{$\pm$0.49} & 76.24\scriptsize{$\pm$0.43} & 42.93\scriptsize{$\pm$0.56} & 77.27\scriptsize{$\pm$0.28} & 40.23\scriptsize{$\pm$0.34} & 77.25\scriptsize{$\pm$0.33} & 36.76\scriptsize{$\pm$0.81} & 78.72\scriptsize{$\pm$0.24} & 44.01\scriptsize{$\pm$0.32} \\

ManyDG & 85.22\scriptsize{$\pm$0.22} & 57.94\scriptsize{$\pm$0.26} & 77.24\scriptsize{$\pm$0.24} & 44.59\scriptsize{$\pm$0.36} & 78.71\scriptsize{$\pm$0.12} & 41.35\scriptsize{$\pm$0.27} & 78.40\scriptsize{$\pm$0.23} & 38.27\scriptsize{$\pm$0.56} & 79.89\scriptsize{$\pm$0.10} & 45.54\scriptsize{$\pm$0.17} \\

\midrule

\rowcolor{gray!20} \method{} & \textbf{87.15\scriptsize{$\pm$0.09}} & \textbf{60.02\scriptsize{$\pm$0.15}} & \textbf{79.88\scriptsize{$\pm$0.09}} & \textbf{47.42\scriptsize{$\pm$0.19}} & \textbf{81.35\scriptsize{$\pm$0.13}} & \textbf{43.92\scriptsize{$\pm$0.14}} & \textbf{81.50\scriptsize{$\pm$0.18}} & \textbf{41.64\scriptsize{$\pm$0.28}} & \textbf{82.47\scriptsize{$\pm$0.04}} & \textbf{48.25\scriptsize{$\pm$0.08}} \\

\bottomrule
\end{tabular}}
\end{table*}

\begin{table}[!t]
\centering
\caption{
Ablation study of \method{}. Top: incremental construction from ERM. Bottom: component removal. Worst: lowest target-condition accuracy on HAR-C. eICU results are averaged over regions.
}
\label{tab:ablation}
\vspace{-.1in}
\resizebox{.7\linewidth}{!}{
\setlength{\tabcolsep}{4pt}
\begin{tabular}{l|cc|cc}
\toprule

\multirow{2}{*}[-0.2em]{\textbf{Variant}} & \multicolumn{2}{c}{\textbf{HAR-C}} & \multicolumn{2}{c}{\textbf{eICU}} \\
\cmidrule(l{3pt}r{3pt}){2-3} \cmidrule(l{3pt}r{3pt}){4-5}
 & Avg. & Worst & AUROC & AUPRC \\

\midrule
\multicolumn{5}{c}{\textbf{Incremental construction}} \\
\midrule

ERM & 66.47\scriptsize{$\pm$0.95} & 54.05\scriptsize{$\pm$4.27} & 79.72\scriptsize{$\pm$0.15} & 44.98\scriptsize{$\pm$0.25} \\

\quad + Complementary pretraining & 72.02\scriptsize{$\pm$0.43} & 56.29\scriptsize{$\pm$3.73} & 81.87\scriptsize{$\pm$0.07} & 47.58\scriptsize{$\pm$0.15} \\

\quad + Sampling variations & 79.58\scriptsize{$\pm$0.17} & 73.85\scriptsize{$\pm$0.18} & 82.17\scriptsize{$\pm$0.07} & 47.91\scriptsize{$\pm$0.15} \\

\rowcolor{gray!20} \quad + Worst-view objective (\method{}) & \textbf{80.57\scriptsize{$\pm$0.19}} & \textbf{74.72\scriptsize{$\pm$0.21}} & \textbf{82.47\scriptsize{$\pm$0.04}} & \textbf{48.25\scriptsize{$\pm$0.08}} \\

\midrule
\multicolumn{5}{c}{\textbf{Component removal from \method{}}} \\
\midrule

w/o $\mathcal{L}_{\text{feat}}$ & 80.02\scriptsize{$\pm$0.14} & 74.36\scriptsize{$\pm$0.25} & 80.74\scriptsize{$\pm$0.07} & 46.38\scriptsize{$\pm$0.06} \\

w/o $\mathcal{L}_{\text{samp}}$ & 80.47\scriptsize{$\pm$0.15} & 74.50\scriptsize{$\pm$1.19} & 81.94\scriptsize{$\pm$0.08} & 47.23\scriptsize{$\pm$0.16} \\

w/o $E_{\text{feat}}$ & 79.41\scriptsize{$\pm$0.16} & 72.88\scriptsize{$\pm$1.27} & 81.19\scriptsize{$\pm$0.05} & 46.79\scriptsize{$\pm$0.13} \\

w/o $E_{\text{samp}}$ & 80.02\scriptsize{$\pm$0.14} & 73.96\scriptsize{$\pm$0.15} & 82.27\scriptsize{$\pm$0.06} & 47.65\scriptsize{$\pm$0.15} \\

\bottomrule
\end{tabular}}
\vspace{-.25in}
\end{table}

\subsection{Ablation study}\label{exp:ablation_study}

To answer RQ2, \Cref{tab:ablation} builds \method{} incrementally from ERM and removes each component from the full model.
Every component improves average performance, but their relative importance depends on the type of distribution shift.
On HAR-C, removing sampling variations ($K=1$) causes the largest degradation, while training on the same variations with an average loss already recovers performance from $72.02\%$ to $79.58\%$.
Using the worst-view objective further improves accuracy to $80.57\%$, showing that exposure to diverse sampling processes provides the dominant source of robustness, with minimax optimization preventing difficult variations from being averaged away.
Removing either pretraining objective or encoder branch produces additional degradation, indicating that the complementary feature-centric and sampling-centric representations provide information beyond augmentation alone.

The relative effects differ on eICU~(\Cref{tab:ablation_eicu}).
Here, complementary pretraining contributes more strongly under naturally occurring cross-region shifts, while sampling variations and worst-view optimization still provide consistent improvements.
Thus, the ablations suggest that the components of \method{} play complementary roles: sampling variations are particularly important when the target observation process changes explicitly, whereas representation pretraining becomes more influential when sampling shifts are entangled with broader real-world variation.
Full ablation results are provided in~\Cref{tab:ablation_full,tab:ablation_eicu}.

\subsection{Further analysis}\label{exp:further_analysis}

\paragraph{Generalization across backbone architectures.}

\begin{wrapfigure}{r}{0.42\textwidth}
\centering
\includegraphics[width=\linewidth]{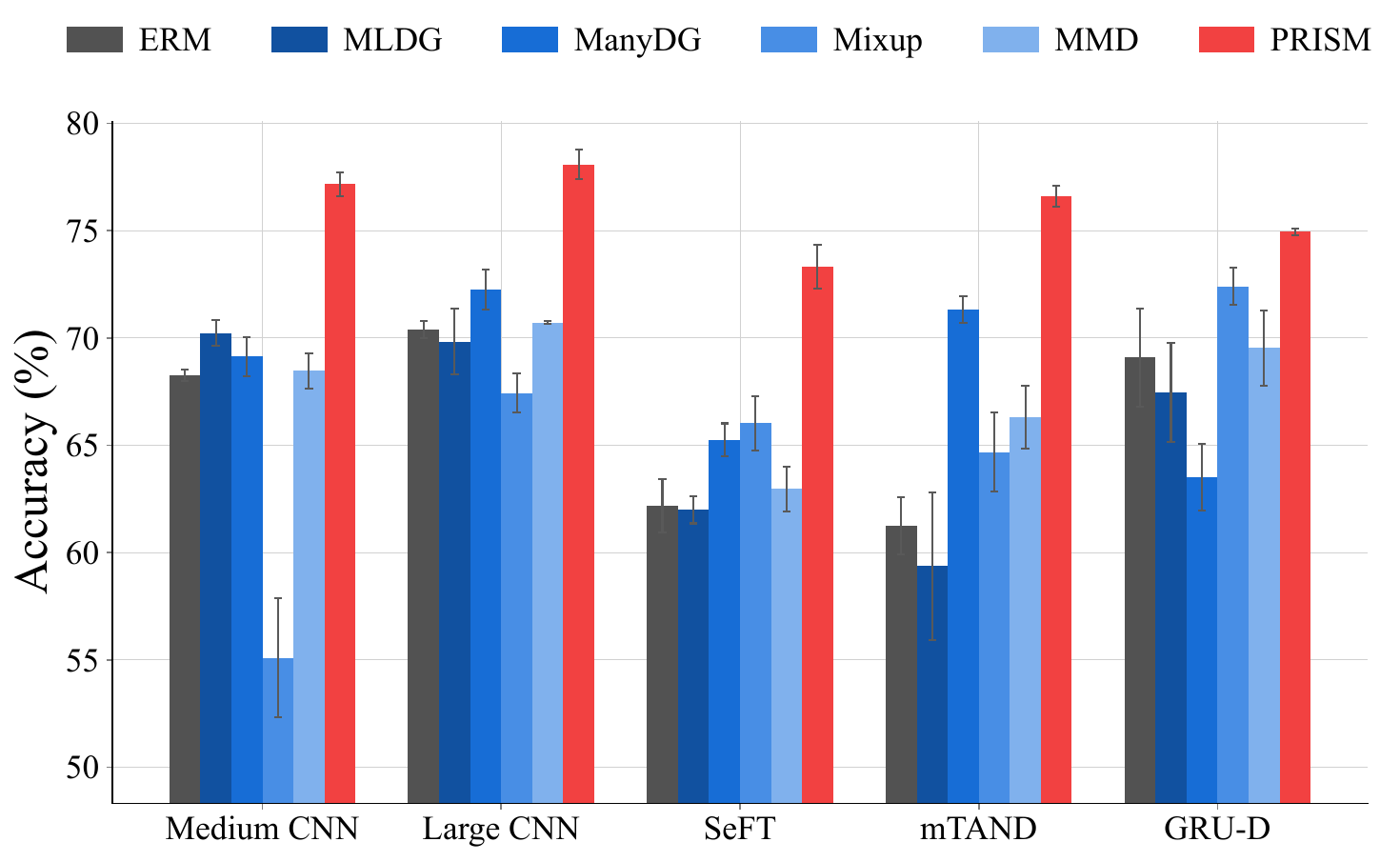}
\vspace{-.2in}
\caption{
Backbone generalization on HAR-C.
We compare \method{} with DG baselines under diverse backbones.
}
\label{fig:backbone_generalization}
\vspace{-.15in}
\end{wrapfigure}

To answer RQ3, we first examine whether the gains of \method{} depend on the CNN architecture used in the main experiments.
As shown in~\Cref{fig:backbone_generalization}, the improvement persists across Medium CNN, Large CNN, SeFT~\citep{seft}, mTAND~\citep{mtand}, and GRU-D~\citep{grud}.

The difference is generally more pronounced under shifted sampling processes than under the source-matched condition, mirroring the pattern observed with the default backbone.
This consistency across convolutional, recurrent, set-based, and attention-based architectures suggests that the robustness of \method{} arises from its training framework rather than a particular architectural inductive bias.
Full results across all HAR-C sampling conditions are provided in~\Cref{sec:additional_experiments}.

\paragraph{Representation analysis.}
We further examine whether the complementary pretraining objectives induce the intended information preferences.
As shown in~\Cref{fig:branch_information}, pretraining makes $E_{\mathrm{feat}}$ more informative about feature values ($R^2$ from 0.01 to 0.09), whereas $E_{\mathrm{samp}}$ becomes highly predictive of sampling patterns (AUROC from 0.73 to 0.97).
Notably, $E_{\mathrm{samp}}$ retains almost no value information despite receiving the full input, indicating that it specializes in the observation structure rather than copying its input.
After downstream supervision, these preferences partially blend, as expected when both representations are jointly optimized for prediction, but the two branches retain complementary tendencies.
This supports the intended role of pretraining as encouraging complementary specialization rather than enforcing rigid disentanglement between feature and sampling information.

\begin{figure*}[!t]
\centering
\includegraphics[width=\textwidth]{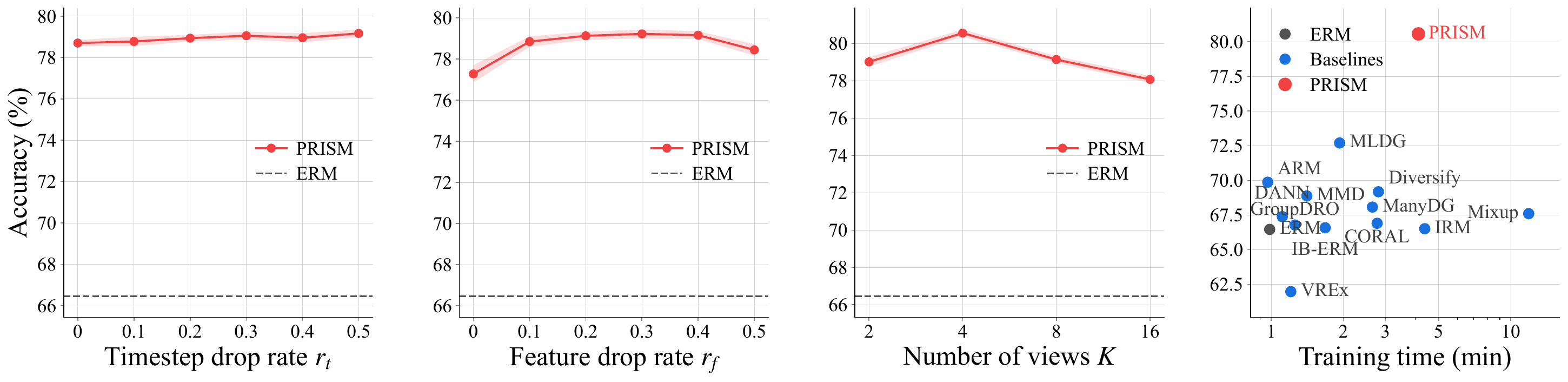}
\vspace{-.2in}
\caption{
Sensitivity and efficiency analyses of \method{}.
From left to right, we vary the timestep drop rate, feature drop rate, and number of sampling views $K$, then compare training time with HAR-C accuracy.
}
\label{fig:ablations}
\vspace{-.2in}
\end{figure*}

\paragraph{Sensitivity and efficiency.}
Finally, \Cref{fig:ablations} examines sensitivity to the augmentation strength and number of sampling variations.
Moderate timestep and feature dropping already provide strong performance, 
while more aggressive perturbations yield diminishing returns.
This indicates that \method{} does not require precise tuning of the augmentation distribution to obtain its robustness gains.
Performance peaks at $K = 4$ and slightly decreases for larger K, possibly because the worst-view objective becomes overly pessimistic; we therefore use $K = 4$ by default.
Although \method{} is more computationally expensive than standard ERM, its higher HAR-C accuracy yields a favorable accuracy--cost trade-off.
Overall, the robustness of \method{} persists across architectures and broad hyperparameter choices without requiring excessive computation.

\section{Conclusion}
This paper studies domain generalization under sampling pattern shifts in irregularly sampled time series.
We introduce a controlled benchmark showing that such shifts degrade performance, induce sampling-specific shortcuts, and challenge existing domain generalization methods.
Motivated by these findings, we propose \method{}, which learns complementary feature-centric and sampling-centric representations and trains robustly over stochastic sampling variations.
Experiments on controlled and real-world benchmarks show improved robustness to unseen sampling shifts, highlighting the need to account for sampling information that is predictive yet unstable across environments.
\section*{AI Use Statement}
Generative AI tools were used to assist with research ideation, code implementation, experimental design and analysis, and manuscript editing.
All AI-assisted outputs were critically reviewed and verified by the authors, who take full responsibility for the methodology, experiments, results, and final content of this work.

\section*{Ethics Statement}
Our experiments use publicly available or de-identified datasets, including the eICU Collaborative Research Database~\citep{eicu}.
The eICU data are accessed and handled in accordance with the corresponding PhysioNet data use requirements, and no restricted patient-level data are shared with third-party services.
Although our work focuses on robustness rather than clinical deployment, sampling patterns in electronic health records may reflect institution-specific patient and acquisition characteristics.
Models exploiting such patterns may therefore inherit environment-specific biases and require careful validation before clinical use.

\section*{Reproducibility Statement}
To facilitate reproducibility, we release the implementation of \method{}, including data preprocessing, HAR-C construction, sampling pattern augmentations, model configurations, training procedures, and evaluation code.
The paper and appendix provide the experimental protocols, hyperparameters, and additional analyses required to reproduce the reported results.
Our code is available at \url{https://anonymous.4open.science/r/\method}.

\bibliography{preamble/reference}
\bibliographystyle{iclr2027_conference}

\clearpage
\appendix
\renewcommand{\thefigure}{A\arabic{figure}} 
\renewcommand{\thetable}{A\arabic{table}} 
\renewcommand{\thetheorem}{A\arabic{theorem}} 
\section{Limitations and Future Work}
Despite its effectiveness, \method{} has several limitations.
Its sampling augmentations should ideally reflect the sampling pattern shifts encountered at test time, but such shifts are generally unknown in advance, making augmentation design challenging.
Future work could learn generators of plausible sampling patterns or adapt the augmentation policy when prior knowledge about likely shifts is available.
Moreover, in real-world settings, changes in the sampling process may be entangled with shifts in the underlying feature distribution, making it difficult to isolate and address sampling shifts alone.
Understanding and handling such coupled shifts remains an important direction for future work.
\section{Related work}
\paragraph{Predictive models for irregular time series.}
The machine learning community has dedicated significant efforts to advancing prediction models for irregular time series, evolving from data-centric to model-centric approaches. Classical imputation methods include simple statistical approaches like mean/zero imputation and forward-filling, and probabilistic methods like Gaussian Process regression~\citep{gp}. These were followed by RNN variants like GRU-D~\citep{grud} and BRITS~\citep{brits} that incorporated learnable interpolation mechanisms; yet, all these approaches often degraded temporal precision and overlooked valuable information in the sampling patterns.
Modern approaches have shifted toward directly processing irregular time series in their native form, with Neural ODEs~\citep{neuralode,latentode} treating them as continuous-time dynamic systems, and set-based architectures like SeFT~\citep{seft}, mTAND~\citep{mtand}, and STraTS~\citep{strats} handling them as observation sets with temporal features. Recent trends show diverse representation paradigms: vision-based approaches that reformulate time series as structured visual inputs~\citep{vitst}, methods leveraging language models for time series tasks~\citep{tan2024language,timellm}, and graph-based frameworks that model sensor dependencies through neural message passing and temporal attention~\citep{raindrop}.
However, despite these advances, most existing approaches have not thoroughly examined their out-of-distribution generalization capabilities, particularly against sampling pattern shifts, critical as real-world applications often encounter significant variations in sampling patterns across different domains.

\paragraph{Modality-agnostic domain generalization.}
Domain generalization aims to learn robust models under distribution shifts by relaxing the \emph{i.i.d. assumption} between training and test distributions, where data from the target domain is unavailable during training.
Existing approaches broadly include data manipulation, representation learning, invariance-based learning, robust optimization, and meta-learning.
Data manipulation methods increase source diversity through augmentations such as Mixup~\citep{mixup}, while representation learning aligns domains through adversarial training (DANN)~\citep{dann}, moment matching (MMD)~\citep{mmd}, or correlation alignment (CORAL)~\citep{coral}.
Methods targeting spurious correlations instead encourage predictors that remain stable across environments, including IRM~\citep{irm}, VREx~\citep{vrex}, and IB\_ERM~\citep{ib_erm}.
Robust optimization approaches such as GroupDRO~\citep{groupdro} optimize worst-group performance, while MLDG~\citep{mldg} and ARM~\citep{arm} use meta-learning to improve generalization across domains.
Despite their broad applicability, these methods are not tailored for sampling pattern shifts as a potentially predictive yet domain-dependent source of shortcut information in irregular time series.

\paragraph{Domain generalization on time series domain.}
Time series domain generalization requires sophisticated approaches to address both the unique characteristics of temporal data and their distribution shift patterns~\citep{woods}. Recent efforts have begun exploring this challenge across various contexts.
WOODS~\citep{woods} introduces comprehensive benchmarks for out-of-distribution generalization across diverse time series modalities, highlighting the unique challenges posed by temporal data.
Diversify~\citep{diversify} approaches time series from a distribution perspective, learning generalized representations by characterizing latent distributions and their dynamic changes over time through adversarial training.
ManyDG~\citep{manydg} addresses patient covariate shifts in healthcare applications by treating each patient as a distinct domain and explicitly removing patient-specific characteristics through orthogonal projection.
However, these methods have achieved only marginal performance improvements in practice~\citep{manydg}, and most do not explicitly consider the sampling irregularity that is inherent in time series sampling patterns, leaving the challenge of domain generalization under varying sampling patterns largely unexplored.
\section{Algorithms}
This section provides the detailed training procedures of \method{}.
Complementary representation pretraining is summarized in~\Cref{alg:pretraining}, followed by sample-wise minimax learning across sampling variations in~\Cref{alg:robust_training}.

\begin{algorithm}[!ht]
\caption{Complementary Representation Pretraining in \method{}}
\label{alg:pretraining}
\begin{algorithmic}[1]
    \Require Source data $\D_s$, encoders $E_{\mathrm{feat}}, E_{\mathrm{samp}}$, decoders $D_{\mathrm{feat}}, D_{\mathrm{samp}}$, augmentation distribution $\mathcal{T}$

    \For{each minibatch $\mathbf{x} \sim \D_s$}
        \State Sample $T \sim \mathcal{T}$ and obtain $\tilde{\mathbf{x}} \gets T(\mathbf{x})$

        \State $\hat{\boldsymbol{v}}
        \gets
        D_{\mathrm{feat}}(E_{\mathrm{feat}}(\tilde{\mathbf{x}}))$

        \State $\mathcal{L}_{\mathrm{feat}}
        \gets
        \mathbb{E}_{\boldsymbol{m}_{l,d}=1,\,\tilde{\boldsymbol{m}}_{l,d}=0}
        \left[
            \ell_{\mathrm{MSE}}
            (\hat{\boldsymbol{v}}_{l,d}, \boldsymbol{v}_{l,d})
        \right]$

        \State $\hat{\boldsymbol{m}}
        \gets
        D_{\mathrm{samp}}(E_{\mathrm{samp}}(\mathbf{x}))$

        \State $\mathcal{L}_{\mathrm{samp}}
        \gets
        \mathbb{E}_{l,d}
        \left[
            \ell_{\mathrm{BCE}}
            (\hat{\boldsymbol{m}}_{l,d}, \boldsymbol{m}_{l,d})
        \right]$

        \State Update $E_{\mathrm{feat}}, E_{\mathrm{samp}}, D_{\mathrm{feat}}, D_{\mathrm{samp}}$
        by minimizing
        $\mathcal{L}_{\mathrm{feat}}+\mathcal{L}_{\mathrm{samp}}$
    \EndFor

    \State Discard $D_{\mathrm{feat}}$ and $D_{\mathrm{samp}}$
    \Ensure Pretrained $E_{\mathrm{feat}}$ and $E_{\mathrm{samp}}$
\end{algorithmic}
\end{algorithm}
\begin{algorithm}[!ht]
\caption{Robust Learning Across Sampling Variations in \method{}}
\label{alg:robust_training}
\begin{algorithmic}[1]
    \Require Labeled source data $\D_s$, pretrained encoders $E_{\mathrm{feat}}, E_{\mathrm{samp}}$, classifier $H$, augmentation distribution $\mathcal{T}$, number of variations $K$

    \State Define
    $F(\mathbf{x})
    \gets
    H([E_{\mathrm{feat}}(\mathbf{x}); E_{\mathrm{samp}}(\mathbf{x})])$

    \For{each minibatch $\{(\mathbf{x}_i, y_i)\}_{i=1}^{B} \sim \D_s$}
        \For{$i = 1, \dots, B$}
            \State Sample $T_{i,1}, \dots, T_{i,K} \sim \mathcal{T}$

            \State $\mathcal{L}_{\mathrm{prism}}^{(i)}
            \gets
            \max_{k \in \{1, \dots, K\}}
            \ell_{\mathrm{CE}}
            \left(
            F(T_{i,k}(\mathbf{x}_i)),
            y_i
            \right)$
        \EndFor

        \State Update $F$ by minimizing
        $\frac{1}{B}
        \sum_{i=1}^{B}
        \mathcal{L}_{\mathrm{prism}}^{(i)}$
    \EndFor

    \Ensure Trained predictor $F$
\end{algorithmic}
\end{algorithm}

\section{Additional Experiments}
\label{sec:additional_experiments}

This section provides additional dataset details, analyses of sampling pattern shifts, and extended evaluations complementing the main experiments.

\subsection{Dataset Details and Additional Analysis}

\Cref{tab:dataset-statistics} summarizes the statistics and evaluation settings of HAR-C, eICU, and PAM.
HAR-C is constructed by subsampling the original regularly observed HAR sequences at a fixed $20\%$ observation rate, whereas eICU and PAM retain their naturally irregular observation patterns.
The three benchmarks provide complementary settings: controlled sampling pattern shifts on HAR-C, naturally occurring cross-region shifts on eICU, and sensor-level sampling shifts on PAM.

\begin{figure*}[!t]
\centering
\includegraphics[width=\textwidth]{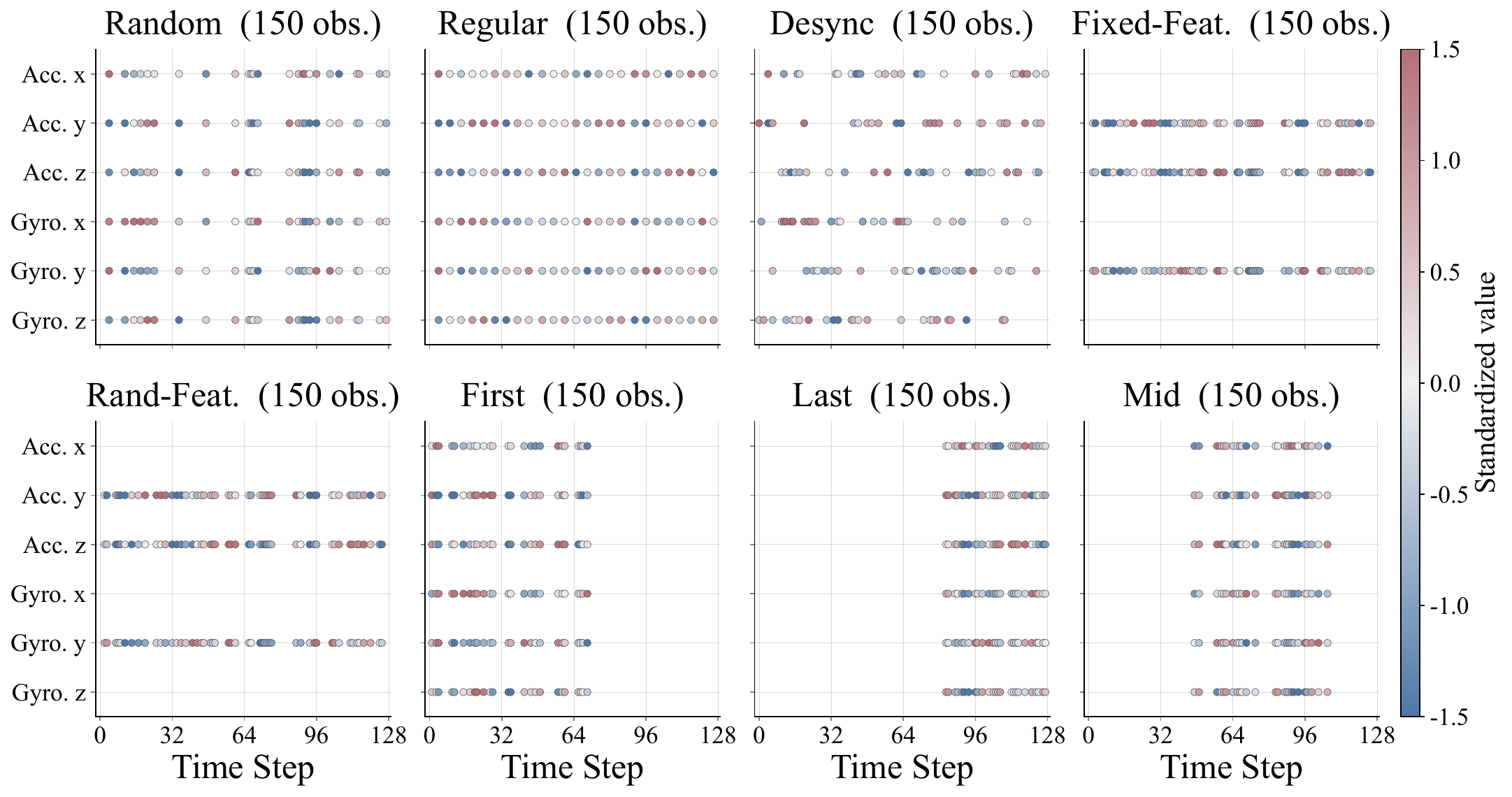}
\vspace{-.15in}
\caption{
Examples of HAR-C observations under diverse sampling pattern shifts.
Each condition retains 150 observations while varying their temporal and feature-wise locations; color indicates the standardized feature value.
}
\label{fig:harc_sample_masks}
\vspace{-.15in}
\end{figure*}

\Cref{fig:harc_sample_masks} visualizes representative HAR-C samples under the eight sampling conditions.
Each condition retains the same number of observations while changing their temporal and feature-wise locations, illustrating how HAR-C isolates changes in the sampling process from changes in the observation budget.

\begin{figure*}[!t]
\centering
\includegraphics[width=\textwidth]{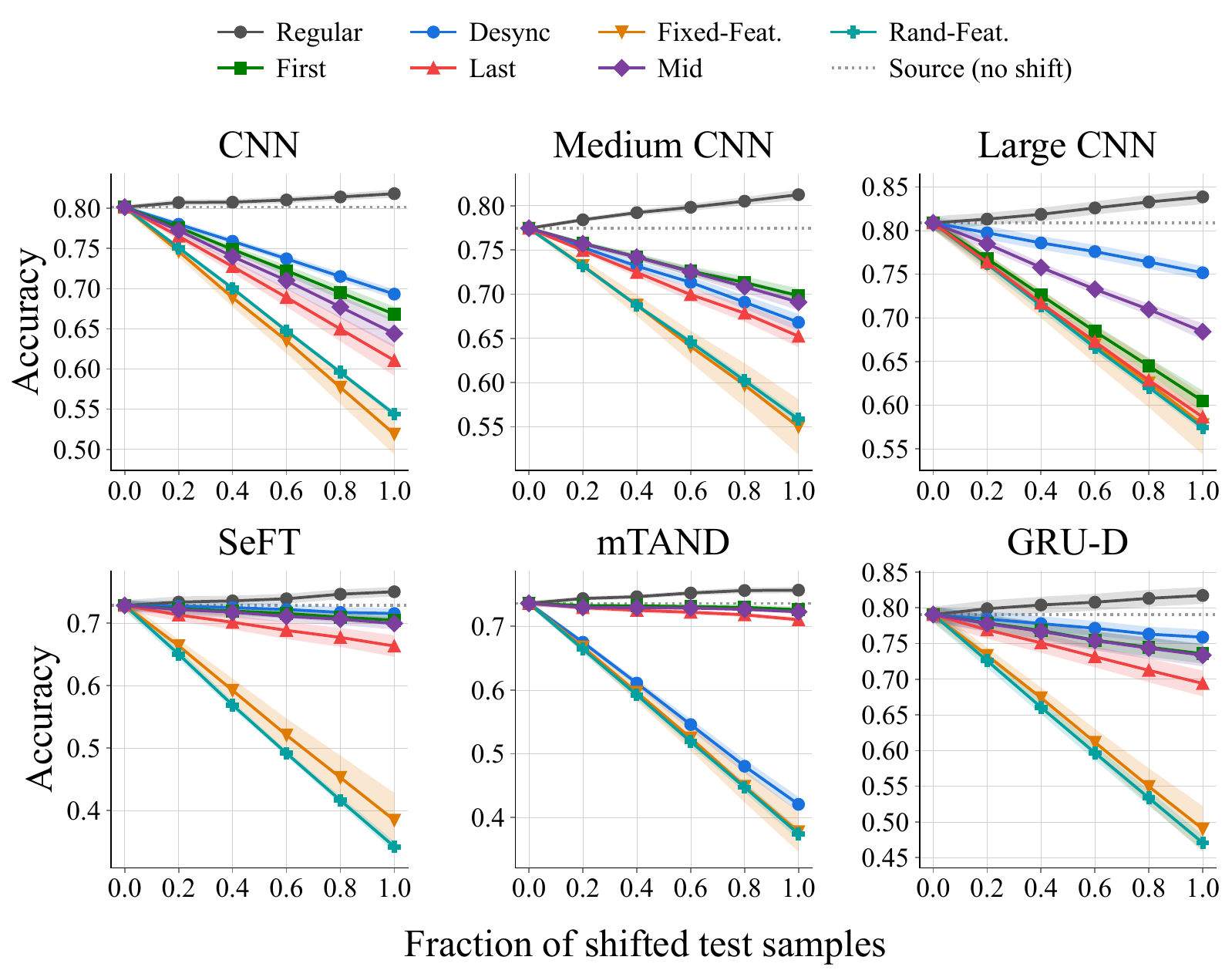}
\vspace{-.15in}
\caption{
Further analysis of sampling pattern shifts across backbone architectures on HAR-C.
We vary the fraction of shifted test samples for CNN, Medium CNN, Large CNN, SeFT, mTAND, and GRU-D.
}
\label{fig:a1_backbones}
\vspace{-.15in}
\end{figure*}

To verify that sensitivity to sampling pattern shifts is not specific to the CNN used in the main analysis, we repeat the shift-severity experiment across Medium CNN, Large CNN, SeFT~\citep{seft}, mTAND~\citep{mtand}, and GRU-D~\citep{grud}.
As shown in~\Cref{fig:a1_backbones}, performance generally deteriorates as the fraction of shifted test samples increases across architectures.

We also report the full sampling-specific shortcut results at $\rho=0.9$ in~\Cref{tab:harc-bias}.
Across DG methods, accuracy is substantially lower when the sampling cue conflicts with the ground-truth label than when it is aligned, complementing the analysis in~\Cref{sec:analysis_sampling}.

\subsection{Additional Evaluation of \method{}}

We further evaluate \method{} on PAM under leave-fixed-sensors-out and across diverse backbone architectures on HAR-C.
On PAM, the fraction of entirely unobserved sensors is varied from $10\%$ to $50\%$, and \method{} remains substantially more robust as the shift becomes more severe.
On HAR-C, the gains of \method{} persist across Medium CNN, Large CNN, SeFT, mTAND, and GRU-D, indicating that its effectiveness is not tied to the default CNN backbone.

We provide the full per-condition ablation results on HAR-C in \Cref{tab:ablation_full} and the per-region results on eICU under leave-one-region-out evaluation in \Cref{tab:ablation_eicu}, complementing the summary in \Cref{tab:ablation}.
Both tables additionally include variants that remove a pretraining objective together with sampling variations.
On eICU, removing any component degrades the average AUROC and AUPRC, with the feature-centric branch showing the largest effect, although the relative ranking of variants can vary across individual target regions.

Finally, we probe the information captured by $E_{\mathrm{feat}}$ and $E_{\mathrm{samp}}$ at random initialization, after pretraining, and after downstream training.
Feature-value information is measured by $R^2$, while sampling-pattern information is measured by AUROC.
The results show that complementary pretraining encourages the two encoders to emphasize different aspects of the input.

\begin{table*}[t]
\centering
\caption{
Dataset statistics for the three benchmarks. Observed fraction is the share of the $T \times F$ lattice that carries a measurement: on HAR-C it is injected by subsampling a fully observed signal at a 20\% rate ($\lfloor 128 \times 0.2 \rfloor = 25$ of 128 steps, synchronised across channels), whereas on eICU and PAM it is the native charting or recording density. eICU is evaluated leave-one-region-out, so its split sizes depend on the held-out region; the Northeast fold is shown, with negative\,/\,positive mortality counts in parentheses. HAR-C holds out 20\% of the training subjects for validation.
}
\label{tab:dataset-statistics}
\vspace{-.05in}
\setlength{\tabcolsep}{6pt}
\resizebox{.8\textwidth}{!}{%
\begin{tabular}{l|ccc}
    \toprule
    & \textbf{HAR-C} & \textbf{eICU} & \textbf{PAM} \\
    \midrule
    Task & Activity recognition & In-hospital mortality & Activity recognition \\
    \# Classes & 6 & 2 & 7 \\
    \# Features $F$ & 6 & 11 & 12 \\
    Sequence length $T$ & 128 & 24 & 50 \\
    Time window & 2.56\,s & 24\,h & --- \\
    Observed fraction & 20\% (25/128 steps) & 38.5\% (native) & 24.9\% (native) \\
    \midrule
    Distribution shift & 8 sampling-pattern shifts & Leave-one-region-out & Leave-sensors-out \\
    \# Test conditions & 8 & 4 regions & 5 severities \\
    \midrule
    Train & 5,881 & 22,543 (19,992 / 2,551) & 4,228 \\
    Val & 1,471 & 5,636 (4,994 / 642) & 1,056 \\
    Test & 2,947 & 2,495 (2,166 / 329) & 1,320 \\
    Total & 10,299 & 30,674 & 6,604 \\
    \bottomrule
\end{tabular}%
}
\end{table*}

\begin{table}[t]
\centering
\caption{
Sampling augmentation settings used for pretraining and downstream robust learning.
}
\label{tab:augmentation_settings}
\begin{tabular}{lccc}
\toprule
Dataset & Pretrain $(r_t, r_f)$ & Downstream $(r_t, r_f)$ & $K$ \\
\midrule
HAR-C
& $(0.5, 0.4)$
& $(0.5,0.4)$
& 4 \\
PAM
& $(0.5, 0.4)$
& $(0.5, 0.4)$
& 4 \\
eICU
& $(0.5, 0)$
&$(0.2, 0)$
& 4 \\
\bottomrule
\end{tabular}
\vspace{-.1in}
\end{table}

\begin{table}[!t]
\centering
\caption{
Sampling-specific shortcut analysis on HAR-C at $\rho=0.9$.
At test time, the sampling cue is bias-aligned, unbiased, or bias-conflicting with the ground-truth label.
}
\label{tab:harc-bias}
\vspace{-.1in}
\resizebox{.6\columnwidth}{!}{
\setlength{\tabcolsep}{4pt}
\begin{tabular}{l|ccc}
\toprule

\textbf{Algorithm} & \textbf{Bias-aligned} & \textbf{Unbiased} & \textbf{Bias-conflicting} \\

\midrule

ERM & 93.99\scriptsize{$\pm$2.44} & 58.00\scriptsize{$\pm$0.69} & 42.59\scriptsize{$\pm$1.86} \\

ARM & 77.51\scriptsize{$\pm$1.30} & 57.61\scriptsize{$\pm$1.35} & 44.36\scriptsize{$\pm$1.56} \\

CORAL & 97.76\scriptsize{$\pm$0.15} & 54.54\scriptsize{$\pm$0.69} & 34.89\scriptsize{$\pm$0.94} \\

DANN & 98.03\scriptsize{$\pm$0.08} & 57.79\scriptsize{$\pm$0.61} & 40.18\scriptsize{$\pm$0.95} \\

GroupDRO & 97.48\scriptsize{$\pm$0.18} & 57.67\scriptsize{$\pm$0.76} & 40.83\scriptsize{$\pm$1.20} \\

IB\_ERM & 93.20\scriptsize{$\pm$2.91} & 54.87\scriptsize{$\pm$0.73} & 36.73\scriptsize{$\pm$1.24} \\

IRM & 96.66\scriptsize{$\pm$2.12} & 56.25\scriptsize{$\pm$0.86} & 39.67\scriptsize{$\pm$1.59} \\

Mixup & 91.23\scriptsize{$\pm$3.14} & 52.17\scriptsize{$\pm$1.09} & 30.29\scriptsize{$\pm$1.72} \\

MLDG & 98.01\scriptsize{$\pm$0.20} & 57.57\scriptsize{$\pm$1.04} & 41.62\scriptsize{$\pm$1.86} \\

MMD & 97.59\scriptsize{$\pm$0.11} & 57.37\scriptsize{$\pm$0.51} & 41.40\scriptsize{$\pm$1.26} \\

VREx & 85.27\scriptsize{$\pm$1.66} & 48.49\scriptsize{$\pm$1.52} & 39.95\scriptsize{$\pm$3.08} \\

Diversify & \textbf{98.09\scriptsize{$\pm$0.09}} & 56.83\scriptsize{$\pm$0.86} & 41.31\scriptsize{$\pm$1.42} \\

ManyDG & 93.55\scriptsize{$\pm$0.79} & 59.17\scriptsize{$\pm$0.81} & 44.87\scriptsize{$\pm$1.26} \\

\midrule

\rowcolor{gray!20} \method{} & 92.29\scriptsize{$\pm$2.59} & \textbf{64.98\scriptsize{$\pm$0.83}} & \textbf{49.11\scriptsize{$\pm$0.74}} \\

\bottomrule
\end{tabular}}
\end{table}

\begin{table*}[!t]
\centering
\caption{
Comparison of \method{} and existing DG methods on PAM under leave-fixed-sensors-out evaluation. A fixed fraction of sensors is entirely unobserved at test time, with the missing-sensor ratio varied from 10\% to 50\%.
}
\label{tab:pam_lso}
\vspace{-.1in}
\resizebox{.8\textwidth}{!}{
\setlength{\tabcolsep}{4pt}
\begin{tabular}{l|ccccc|c}
\toprule

\textbf{Algorithm} & \textbf{10\%} & \textbf{20\%} & \textbf{30\%} & \textbf{40\%} & \textbf{50\%} & \textbf{Avg.} \\

\midrule

ERM & 85.46\scriptsize{$\pm$0.47} & 79.67\scriptsize{$\pm$1.54} & 67.82\scriptsize{$\pm$3.54} & 57.24\scriptsize{$\pm$4.85} & 50.33\scriptsize{$\pm$4.66} & 68.11\scriptsize{$\pm$2.39} \\

ARM & 85.63\scriptsize{$\pm$0.43} & 80.10\scriptsize{$\pm$1.75} & 69.86\scriptsize{$\pm$3.96} & 59.39\scriptsize{$\pm$4.45} & 53.35\scriptsize{$\pm$5.04} & 69.66\scriptsize{$\pm$2.55} \\

CORAL & 85.85\scriptsize{$\pm$0.45} & 80.51\scriptsize{$\pm$1.73} & 69.01\scriptsize{$\pm$3.92} & 59.70\scriptsize{$\pm$4.82} & 53.87\scriptsize{$\pm$4.90} & 69.79\scriptsize{$\pm$2.69} \\

DANN & 85.34\scriptsize{$\pm$0.46} & 79.51\scriptsize{$\pm$1.73} & 69.01\scriptsize{$\pm$3.68} & 57.30\scriptsize{$\pm$4.23} & 52.03\scriptsize{$\pm$4.65} & 68.64\scriptsize{$\pm$2.31} \\

GroupDRO & 85.65\scriptsize{$\pm$0.32} & 79.47\scriptsize{$\pm$1.59} & 67.52\scriptsize{$\pm$3.23} & 57.31\scriptsize{$\pm$3.93} & 51.26\scriptsize{$\pm$4.15} & 68.24\scriptsize{$\pm$2.16} \\

IB\_ERM & 85.66\scriptsize{$\pm$0.37} & 80.49\scriptsize{$\pm$1.44} & 68.58\scriptsize{$\pm$3.67} & 58.78\scriptsize{$\pm$4.34} & 52.14\scriptsize{$\pm$4.40} & 69.13\scriptsize{$\pm$2.33} \\

IRM & 85.71\scriptsize{$\pm$0.45} & 79.47\scriptsize{$\pm$1.70} & 67.52\scriptsize{$\pm$3.85} & 58.29\scriptsize{$\pm$4.17} & 52.28\scriptsize{$\pm$4.08} & 68.65\scriptsize{$\pm$2.30} \\

Mixup & \textbf{86.18\scriptsize{$\pm$0.32}} & 81.61\scriptsize{$\pm$1.42} & 72.55\scriptsize{$\pm$3.20} & 65.19\scriptsize{$\pm$3.70} & 58.99\scriptsize{$\pm$4.54} & 72.90\scriptsize{$\pm$2.24} \\

MLDG & 85.82\scriptsize{$\pm$0.51} & 81.15\scriptsize{$\pm$1.35} & 71.06\scriptsize{$\pm$3.86} & 61.26\scriptsize{$\pm$4.35} & 54.22\scriptsize{$\pm$4.57} & 70.70\scriptsize{$\pm$2.33} \\

MMD & 85.93\scriptsize{$\pm$0.51} & 80.57\scriptsize{$\pm$1.78} & 67.92\scriptsize{$\pm$3.80} & 58.08\scriptsize{$\pm$4.85} & 51.54\scriptsize{$\pm$4.75} & 68.81\scriptsize{$\pm$2.55} \\

VREx & 85.89\scriptsize{$\pm$0.38} & 80.32\scriptsize{$\pm$1.44} & 69.47\scriptsize{$\pm$3.26} & 60.64\scriptsize{$\pm$3.84} & 54.14\scriptsize{$\pm$3.90} & 70.09\scriptsize{$\pm$2.07} \\

Diversify & 85.47\scriptsize{$\pm$0.36} & 80.20\scriptsize{$\pm$1.76} & 67.63\scriptsize{$\pm$3.05} & 58.01\scriptsize{$\pm$4.36} & 52.33\scriptsize{$\pm$4.61} & 68.73\scriptsize{$\pm$2.18} \\

ManyDG & 85.08\scriptsize{$\pm$0.57} & 80.04\scriptsize{$\pm$1.29} & 68.82\scriptsize{$\pm$3.52} & 59.52\scriptsize{$\pm$4.44} & 52.27\scriptsize{$\pm$4.85} & 69.15\scriptsize{$\pm$2.47} \\

\midrule

\rowcolor{gray!20} \method{} & 85.48\scriptsize{$\pm$0.28} & \textbf{85.36\scriptsize{$\pm$0.31}} & \textbf{84.42\scriptsize{$\pm$0.49}} & \textbf{83.25\scriptsize{$\pm$1.05}} & \textbf{81.97\scriptsize{$\pm$1.01}} & \textbf{84.10\scriptsize{$\pm$0.48}} \\

\bottomrule
\end{tabular}}
\end{table*}

\begin{table*}[!t]
\centering
\caption{
Ablation study of \method{} on HAR-C.
We ablate $\mathcal{L}_{\mathrm{feat}}$, $\mathcal{L}_{\mathrm{samp}}$, $E_{\mathrm{feat}}$, $E_{\mathrm{samp}}$, and the sampling-variation and worst-view components of downstream training.
}
\label{tab:ablation_full}
\vspace{-.1in}
\resizebox{\textwidth}{!}{
\setlength{\tabcolsep}{4pt}
\begin{tabular}{l|cccccccc|c}
    \toprule

\textbf{Variant} & \textbf{Random} & \textbf{Regular} & \textbf{Desync} & \textbf{Fixed-Feat.} & \textbf{Rand-Feat.} & \textbf{First} & \textbf{Last} & \textbf{Mid} & \textbf{Avg.} \\

    \midrule

    ERM & 79.92\scriptsize{$\pm$0.55} & 80.97\scriptsize{$\pm$0.81} & 68.91\scriptsize{$\pm$0.98} & 54.05\scriptsize{$\pm$4.27} & 54.48\scriptsize{$\pm$0.26} & 67.25\scriptsize{$\pm$1.52} & 61.10\scriptsize{$\pm$3.20} & 65.06\scriptsize{$\pm$2.50} & 66.47\scriptsize{$\pm$0.95} \\

    w/o $\mathcal{L}_{\mathrm{feat}}$ \& w/o sampling variations & 81.12\scriptsize{$\pm$0.39} & 83.47\scriptsize{$\pm$0.42} & 73.29\scriptsize{$\pm$0.90} & 55.18\scriptsize{$\pm$3.88} & 57.09\scriptsize{$\pm$0.63} & 73.36\scriptsize{$\pm$1.00} & 70.89\scriptsize{$\pm$2.31} & 72.31\scriptsize{$\pm$1.50} & 70.84\scriptsize{$\pm$0.68} \\

    w/o $\mathcal{L}_{\mathrm{samp}}$ \& w/o sampling variations & 82.86\scriptsize{$\pm$0.25} & 85.61\scriptsize{$\pm$0.33} & 78.93\scriptsize{$\pm$0.55} & 58.59\scriptsize{$\pm$3.49} & 58.32\scriptsize{$\pm$0.57} & 75.47\scriptsize{$\pm$0.62} & 66.17\scriptsize{$\pm$1.44} & 70.67\scriptsize{$\pm$0.91} & 72.08\scriptsize{$\pm$0.70} \\

    \midrule

    \multicolumn{10}{c}{\textbf{Pretraining objective}} \\

    \midrule

    w/o $\mathcal{L}_{\mathrm{feat}}$ & 83.47\scriptsize{$\pm$0.20} & 85.73\scriptsize{$\pm$0.15} & 82.50\scriptsize{$\pm$0.28} & {74.60\scriptsize{$\pm$1.15}} & 74.36\scriptsize{$\pm$0.25} & 80.66\scriptsize{$\pm$0.20} & 78.87\scriptsize{$\pm$0.23} & 79.98\scriptsize{$\pm$0.29} & 80.02\scriptsize{$\pm$0.14} \\

    w/o $\mathcal{L}_{\mathrm{samp}}$ & \textbf{84.08\scriptsize{$\pm$0.13}} & {86.31\scriptsize{$\pm$0.21}} & \textbf{83.07\scriptsize{$\pm$0.16}} & 74.50\scriptsize{$\pm$1.19} & {74.68\scriptsize{$\pm$0.18}} & 81.01\scriptsize{$\pm$0.14} & \textbf{79.27\scriptsize{$\pm$0.24}} & {80.84\scriptsize{$\pm$0.23}} & {80.47\scriptsize{$\pm$0.15}} \\

    \midrule

    \multicolumn{10}{c}{\textbf{Encoder}} \\

    \midrule

    w/o $E_{\mathrm{feat}}$ & 83.08\scriptsize{$\pm$0.17} & 84.96\scriptsize{$\pm$0.21} & 82.05\scriptsize{$\pm$0.23} & 72.88\scriptsize{$\pm$1.27} & 73.69\scriptsize{$\pm$0.16} & 80.31\scriptsize{$\pm$0.19} & 78.20\scriptsize{$\pm$0.24} & 80.09\scriptsize{$\pm$0.24} & 79.41\scriptsize{$\pm$0.16} \\

    w/o $E_{\mathrm{samp}}$ & 83.47\scriptsize{$\pm$0.20} & 85.58\scriptsize{$\pm$0.22} & 82.68\scriptsize{$\pm$0.20} & 74.18\scriptsize{$\pm$1.12} & 73.96\scriptsize{$\pm$0.15} & {81.12\scriptsize{$\pm$0.26}} & 78.62\scriptsize{$\pm$0.23} & 80.54\scriptsize{$\pm$0.39} & 80.02\scriptsize{$\pm$0.14} \\

    \midrule

    \multicolumn{10}{c}{\textbf{Downstream objective}} \\

    \midrule

    w/o sampling variations ($K{=}1$) & 82.08\scriptsize{$\pm$0.34} & 83.97\scriptsize{$\pm$0.66} & 76.73\scriptsize{$\pm$0.62} & 56.29\scriptsize{$\pm$3.73} & 58.35\scriptsize{$\pm$0.41} & 75.71\scriptsize{$\pm$0.66} & 70.60\scriptsize{$\pm$0.96} & 72.46\scriptsize{$\pm$1.13} & 72.02\scriptsize{$\pm$0.43} \\

    Average instead of worst-view loss & 82.96\scriptsize{$\pm$0.23} & 85.69\scriptsize{$\pm$0.18} & 81.33\scriptsize{$\pm$0.21} & 73.94\scriptsize{$\pm$1.36} & 73.85\scriptsize{$\pm$0.18} & 80.55\scriptsize{$\pm$0.23} & 78.66\scriptsize{$\pm$0.41} & 79.66\scriptsize{$\pm$0.21} & 79.58\scriptsize{$\pm$0.17} \\

    \midrule

    \cellcolor{gray!20}{\method{}} & \cellcolor{gray!20}{\textbf{84.08\scriptsize{$\pm$0.16}}} & \cellcolor{gray!20}{\textbf{86.49\scriptsize{$\pm$0.21}}} & \cellcolor{gray!20}{{82.81\scriptsize{$\pm$0.27}}} & \cellcolor{gray!20}{\textbf{74.90\scriptsize{$\pm$1.29}}} & \cellcolor{gray!20}{\textbf{74.72\scriptsize{$\pm$0.21}}} & \cellcolor{gray!20}{\textbf{81.28\scriptsize{$\pm$0.20}}} & \cellcolor{gray!20}{{79.22\scriptsize{$\pm$0.35}}} & \cellcolor{gray!20}{\textbf{81.02\scriptsize{$\pm$0.23}}} & \cellcolor{gray!20}{\textbf{80.57\scriptsize{$\pm$0.19}}} \\

    \bottomrule
\end{tabular}}
\vspace{-.2in}
\end{table*}

\begin{table*}[!t]
\centering
\caption{
Ablation study of \method{} on eICU under leave-one-region-out evaluation.
We evaluate the same pretraining, encoder, and downstream-objective ablations as on HAR-C across the four target regions.
}
\label{tab:ablation_eicu}
\vspace{-.1in}
\resizebox{\textwidth}{!}{
\setlength{\tabcolsep}{4pt}
\begin{tabular}{l|cc|cc|cc|cc|cc}
\toprule

\multirow{2}{*}[-0.2em]{\textbf{Variant}} & \multicolumn{2}{c}{\textbf{Northeast}} & \multicolumn{2}{c}{\textbf{West}} & \multicolumn{2}{c}{\textbf{South}} & \multicolumn{2}{c}{\textbf{Midwest}} & \multicolumn{2}{c}{\textbf{Avg.}} \\
\cmidrule(l{3pt}r{3pt}){2-3} \cmidrule(l{3pt}r{3pt}){4-5} \cmidrule(l{3pt}r{3pt}){6-7} \cmidrule(l{3pt}r{3pt}){8-9} \cmidrule(l{3pt}r{3pt}){10-11}
 & AUROC & AUPRC & AUROC & AUPRC & AUROC & AUPRC & AUROC & AUPRC & AUROC & AUPRC \\

\midrule

ERM~\citep{erm} & 85.30\scriptsize{$\pm$0.20} & 57.91\scriptsize{$\pm$0.35} & 76.89\scriptsize{$\pm$0.20} & 43.77\scriptsize{$\pm$0.19} & 78.59\scriptsize{$\pm$0.20} & 40.96\scriptsize{$\pm$0.28} & 78.09\scriptsize{$\pm$0.34} & 37.27\scriptsize{$\pm$0.98} & 79.72\scriptsize{$\pm$0.15} & 44.98\scriptsize{$\pm$0.25} \\
w/o $\mathcal{L}_{\mathrm{feat}}$ \& w/o sampling variations & 83.75\scriptsize{$\pm$0.25} & 55.65\scriptsize{$\pm$0.52} & 77.94\scriptsize{$\pm$0.16} & 44.70\scriptsize{$\pm$0.26} & 76.71\scriptsize{$\pm$0.21} & 38.52\scriptsize{$\pm$0.30} & 76.89\scriptsize{$\pm$0.21} & 35.66\scriptsize{$\pm$0.47} & 78.82\scriptsize{$\pm$0.07} & 43.63\scriptsize{$\pm$0.15} \\
w/o $\mathcal{L}_{\mathrm{samp}}$ \& w/o sampling variations & 86.04\scriptsize{$\pm$0.10} & 58.48\scriptsize{$\pm$0.24} & 78.12\scriptsize{$\pm$0.11} & 44.86\scriptsize{$\pm$0.25} & 79.41\scriptsize{$\pm$0.18} & 41.41\scriptsize{$\pm$0.32} & 79.23\scriptsize{$\pm$0.15} & 37.85\scriptsize{$\pm$0.34} & 80.70\scriptsize{$\pm$0.06} & 45.65\scriptsize{$\pm$0.15} \\
\midrule
\multicolumn{11}{c}{\textbf{Pretraining objective}} \\
\midrule
w/o $\mathcal{L}_{\mathrm{feat}}$ & 85.08\scriptsize{$\pm$0.09} & 57.62\scriptsize{$\pm$0.27} & 78.69\scriptsize{$\pm$0.19} & 45.61\scriptsize{$\pm$0.25} & 79.42\scriptsize{$\pm$0.08} & 42.05\scriptsize{$\pm$0.12} & 79.76\scriptsize{$\pm$0.16} & 40.25\scriptsize{$\pm$0.20} & 80.74\scriptsize{$\pm$0.07} & 46.38\scriptsize{$\pm$0.06} \\
w/o $\mathcal{L}_{\mathrm{samp}}$ & 86.71\scriptsize{$\pm$0.13} & 58.96\scriptsize{$\pm$0.19} & 78.92\scriptsize{$\pm$0.22} & 45.68\scriptsize{$\pm$0.40} & 80.91\scriptsize{$\pm$0.18} & 43.38\scriptsize{$\pm$0.24} & 81.21\scriptsize{$\pm$0.09} & 40.89\scriptsize{$\pm$0.24} & 81.94\scriptsize{$\pm$0.08} & 47.23\scriptsize{$\pm$0.16} \\
\midrule
\multicolumn{11}{c}{\textbf{Encoder}} \\
\midrule
w/o $E_{\mathrm{feat}}$ & 85.83\scriptsize{$\pm$0.13} & 58.15\scriptsize{$\pm$0.20} & 79.21\scriptsize{$\pm$0.14} & 46.28\scriptsize{$\pm$0.24} & 79.70\scriptsize{$\pm$0.10} & 42.29\scriptsize{$\pm$0.16} & 80.00\scriptsize{$\pm$0.19} & 40.42\scriptsize{$\pm$0.27} & 81.19\scriptsize{$\pm$0.05} & 46.79\scriptsize{$\pm$0.13} \\
w/o $E_{\mathrm{samp}}$ & 86.99\scriptsize{$\pm$0.07} & 59.37\scriptsize{$\pm$0.21} & 79.20\scriptsize{$\pm$0.13} & 46.47\scriptsize{$\pm$0.15} & 81.34\scriptsize{$\pm$0.08} & 43.61\scriptsize{$\pm$0.23} & \textbf{81.56\scriptsize{$\pm$0.19}} & 41.15\scriptsize{$\pm$0.39} & 82.27\scriptsize{$\pm$0.06} & 47.65\scriptsize{$\pm$0.15} \\
\midrule
\multicolumn{11}{c}{\textbf{Downstream objective}} \\
\midrule
w/o sampling variations ($K{=}1$) & 87.00\scriptsize{$\pm$0.14} & \textbf{60.48\scriptsize{$\pm$0.33}} & 79.85\scriptsize{$\pm$0.16} & 47.40\scriptsize{$\pm$0.18} & 80.15\scriptsize{$\pm$0.12} & 42.43\scriptsize{$\pm$0.23} & 80.47\scriptsize{$\pm$0.12} & 39.99\scriptsize{$\pm$0.24} & 81.87\scriptsize{$\pm$0.07} & 47.58\scriptsize{$\pm$0.15} \\
Average instead of worst-view loss & 86.92\scriptsize{$\pm$0.13} & 60.35\scriptsize{$\pm$0.28} & \textbf{80.21\scriptsize{$\pm$0.22}} & \textbf{47.77\scriptsize{$\pm$0.29}} & 80.63\scriptsize{$\pm$0.08} & 43.09\scriptsize{$\pm$0.23} & 80.91\scriptsize{$\pm$0.10} & 40.43\scriptsize{$\pm$0.29} & 82.17\scriptsize{$\pm$0.07} & 47.91\scriptsize{$\pm$0.15} \\
\midrule
\rowcolor{gray!20} \method{} & \textbf{87.15\scriptsize{$\pm$0.09}} & 60.02\scriptsize{$\pm$0.15} & 79.88\scriptsize{$\pm$0.09} & 47.42\scriptsize{$\pm$0.19} & \textbf{81.35\scriptsize{$\pm$0.13}} & \textbf{43.92\scriptsize{$\pm$0.14}} & 81.50\scriptsize{$\pm$0.18} & \textbf{41.64\scriptsize{$\pm$0.28}} & \textbf{82.47\scriptsize{$\pm$0.04}} & \textbf{48.25\scriptsize{$\pm$0.08}} \\
\bottomrule
\end{tabular}
}
\end{table*}

\begin{table*}[!ht]
\centering
\caption{
Architecture generalization on HAR-C.
We compare \method{} and DG baselines across Medium CNN, Large CNN, SeFT, mTAND, and GRU-D under the eight sampling conditions.
}
\label{tab:backbone}
\vspace{-.1in}
\resizebox{\textwidth}{!}{
\setlength{\tabcolsep}{4pt}
\begin{tabular}{l|cccccccc|c}
    \toprule

\textbf{Algorithm} & \textbf{Random} & \textbf{Regular} & \textbf{Desync} & \textbf{Fixed-Feat.} & \textbf{Rand-Feat.} & \textbf{First} & \textbf{Last} & \textbf{Mid} & \textbf{Avg.} \\

    \midrule

    \multicolumn{10}{c}{\textbf{Medium CNN}} \\
    \midrule

    ERM & 77.78\scriptsize{$\pm$0.17} & 81.25\scriptsize{$\pm$0.57} & 66.81\scriptsize{$\pm$0.95} & 54.98\scriptsize{$\pm$3.12} & {55.85\scriptsize{$\pm$0.48}} & 69.82\scriptsize{$\pm$0.81} & 65.24\scriptsize{$\pm$1.24} & 69.09\scriptsize{$\pm$0.91} & 67.60\scriptsize{$\pm$0.38} \\

    MLDG & {80.48\scriptsize{$\pm$0.21}} & {84.03\scriptsize{$\pm$0.29}} & {71.75\scriptsize{$\pm$0.91}} & 56.36\scriptsize{$\pm$2.86} & 55.55\scriptsize{$\pm$0.50} & {71.67\scriptsize{$\pm$1.28}} & {67.01\scriptsize{$\pm$0.97}} & {72.34\scriptsize{$\pm$0.54}} & {69.90\scriptsize{$\pm$0.43}} \\

    ManyDG & 78.92\scriptsize{$\pm$0.45} & 82.90\scriptsize{$\pm$0.47} & 70.24\scriptsize{$\pm$0.62} & 52.69\scriptsize{$\pm$2.95} & 55.00\scriptsize{$\pm$0.61} & 71.12\scriptsize{$\pm$1.09} & 63.35\scriptsize{$\pm$1.41} & 69.44\scriptsize{$\pm$1.05} & 67.96\scriptsize{$\pm$0.75} \\

    Mixup & 79.61\scriptsize{$\pm$0.32} & 59.91\scriptsize{$\pm$4.40} & 45.72\scriptsize{$\pm$2.87} & 45.48\scriptsize{$\pm$3.00} & 42.73\scriptsize{$\pm$2.06} & 60.02\scriptsize{$\pm$1.67} & 40.90\scriptsize{$\pm$2.61} & 53.13\scriptsize{$\pm$2.42} & 53.44\scriptsize{$\pm$1.87} \\

    MMD & 77.94\scriptsize{$\pm$0.33} & 80.97\scriptsize{$\pm$0.57} & 67.86\scriptsize{$\pm$0.63} & {56.98\scriptsize{$\pm$2.78}} & 55.80\scriptsize{$\pm$0.44} & 69.27\scriptsize{$\pm$0.83} & 63.39\scriptsize{$\pm$1.08} & 68.21\scriptsize{$\pm$0.89} & 67.55\scriptsize{$\pm$0.33} \\

    \cellcolor{gray!20}{\method{}} & \cellcolor{gray!20}{\textbf{82.79\scriptsize{$\pm$0.17}}} & \cellcolor{gray!20}{\textbf{84.26\scriptsize{$\pm$0.17}}} & \cellcolor{gray!20}{\textbf{79.48\scriptsize{$\pm$0.27}}} & \cellcolor{gray!20}{\textbf{74.97\scriptsize{$\pm$0.82}}} & \cellcolor{gray!20}{\textbf{74.55\scriptsize{$\pm$0.45}}} & \cellcolor{gray!20}{\textbf{80.45\scriptsize{$\pm$0.33}}} & \cellcolor{gray!20}{\textbf{77.36\scriptsize{$\pm$0.36}}} & \cellcolor{gray!20}{\textbf{79.50\scriptsize{$\pm$0.35}}} & \cellcolor{gray!20}{\textbf{79.17\scriptsize{$\pm$0.18}}} \\

    \midrule

    \multicolumn{10}{c}{\textbf{Large CNN}} \\
    \midrule

    ERM & 80.71\scriptsize{$\pm$0.80} & 83.86\scriptsize{$\pm$0.83} & 75.17\scriptsize{$\pm$0.77} & 57.76\scriptsize{$\pm$3.41} & 57.42\scriptsize{$\pm$0.46} & 60.46\scriptsize{$\pm$1.22} & 58.64\scriptsize{$\pm$0.75} & 68.42\scriptsize{$\pm$0.92} & 67.80\scriptsize{$\pm$0.60} \\

    MLDG & \textbf{83.43\scriptsize{$\pm$0.51}} & \textbf{85.44\scriptsize{$\pm$0.37}} & {77.46\scriptsize{$\pm$0.72}} & 57.67\scriptsize{$\pm$2.95} & {57.77\scriptsize{$\pm$0.48}} & 62.84\scriptsize{$\pm$2.53} & 52.69\scriptsize{$\pm$2.85} & 66.04\scriptsize{$\pm$2.09} & 67.92\scriptsize{$\pm$0.92} \\

    ManyDG & 82.21\scriptsize{$\pm$0.38} & {84.89\scriptsize{$\pm$0.59}} & 76.54\scriptsize{$\pm$0.41} & 55.58\scriptsize{$\pm$3.56} & 56.95\scriptsize{$\pm$0.35} & {74.51\scriptsize{$\pm$0.96}} & {68.80\scriptsize{$\pm$2.25}} & {73.95\scriptsize{$\pm$0.96}} & {71.68\scriptsize{$\pm$0.60}} \\

    Mixup & 83.04\scriptsize{$\pm$0.36} & 74.84\scriptsize{$\pm$2.73} & 66.86\scriptsize{$\pm$2.04} & 58.42\scriptsize{$\pm$3.28} & 55.16\scriptsize{$\pm$0.93} & 60.84\scriptsize{$\pm$1.46} & 46.39\scriptsize{$\pm$1.16} & 64.22\scriptsize{$\pm$1.39} & 63.72\scriptsize{$\pm$0.95} \\

    MMD & 81.70\scriptsize{$\pm$0.19} & 84.52\scriptsize{$\pm$0.44} & 75.53\scriptsize{$\pm$0.65} & {58.85\scriptsize{$\pm$2.45}} & 57.55\scriptsize{$\pm$0.48} & 60.75\scriptsize{$\pm$1.70} & 60.54\scriptsize{$\pm$2.18} & 69.04\scriptsize{$\pm$1.93} & 68.56\scriptsize{$\pm$0.70} \\

    \cellcolor{gray!20}{\method{}} & \cellcolor{gray!20}{{83.26\scriptsize{$\pm$0.17}}} & \cellcolor{gray!20}{84.84\scriptsize{$\pm$0.20}} & \cellcolor{gray!20}{\textbf{81.60\scriptsize{$\pm$0.37}}} & \cellcolor{gray!20}{\textbf{76.15\scriptsize{$\pm$0.78}}} & \cellcolor{gray!20}{\textbf{75.16\scriptsize{$\pm$0.32}}} & \cellcolor{gray!20}{\textbf{80.50\scriptsize{$\pm$0.34}}} & \cellcolor{gray!20}{\textbf{77.56\scriptsize{$\pm$0.40}}} & \cellcolor{gray!20}{\textbf{80.24\scriptsize{$\pm$0.35}}} & \cellcolor{gray!20}{\textbf{79.92\scriptsize{$\pm$0.18}}} \\

    \midrule

    \multicolumn{10}{c}{\textbf{SeFT}} \\
    \midrule

    ERM & 72.87\scriptsize{$\pm$0.78} & 75.04\scriptsize{$\pm$0.75} & 71.55\scriptsize{$\pm$0.82} & 38.43\scriptsize{$\pm$4.37} & 34.18\scriptsize{$\pm$0.83} & 70.48\scriptsize{$\pm$0.80} & 66.34\scriptsize{$\pm$1.72} & 69.96\scriptsize{$\pm$0.94} & 62.36\scriptsize{$\pm$0.85} \\

    MLDG & 71.27\scriptsize{$\pm$1.17} & 73.26\scriptsize{$\pm$1.05} & 70.65\scriptsize{$\pm$1.20} & 39.80\scriptsize{$\pm$2.88} & 39.60\scriptsize{$\pm$1.09} & 69.53\scriptsize{$\pm$0.98} & 67.79\scriptsize{$\pm$1.36} & 69.04\scriptsize{$\pm$1.15} & 62.62\scriptsize{$\pm$0.90} \\

    ManyDG & 72.63\scriptsize{$\pm$0.57} & 74.67\scriptsize{$\pm$0.63} & 72.01\scriptsize{$\pm$0.61} & {52.90\scriptsize{$\pm$2.48}} & {44.60\scriptsize{$\pm$0.84}} & 69.51\scriptsize{$\pm$0.55} & 68.05\scriptsize{$\pm$0.82} & 69.81\scriptsize{$\pm$0.63} & 65.52\scriptsize{$\pm$0.43} \\

    Mixup & {75.82\scriptsize{$\pm$0.58}} & \textbf{77.93\scriptsize{$\pm$0.43}} & {75.28\scriptsize{$\pm$0.35}} & 44.73\scriptsize{$\pm$4.73} & 42.84\scriptsize{$\pm$1.27} & {72.58\scriptsize{$\pm$0.83}} & {70.47\scriptsize{$\pm$1.04}} & {73.15\scriptsize{$\pm$0.54}} & {66.60\scriptsize{$\pm$0.79}} \\

    MMD & 72.68\scriptsize{$\pm$0.77} & 74.51\scriptsize{$\pm$0.67} & 71.75\scriptsize{$\pm$0.63} & 37.29\scriptsize{$\pm$4.90} & 35.66\scriptsize{$\pm$1.16} & 70.43\scriptsize{$\pm$0.69} & 66.69\scriptsize{$\pm$1.02} & 69.17\scriptsize{$\pm$0.77} & 62.27\scriptsize{$\pm$0.68} \\

    \cellcolor{gray!20}{\method{}} & \cellcolor{gray!20}{\textbf{76.15\scriptsize{$\pm$1.71}}} & \cellcolor{gray!20}{{76.98\scriptsize{$\pm$1.71}}} & \cellcolor{gray!20}{\textbf{76.22\scriptsize{$\pm$1.72}}} & \cellcolor{gray!20}{\textbf{68.80\scriptsize{$\pm$1.42}}} & \cellcolor{gray!20}{\textbf{66.64\scriptsize{$\pm$1.65}}} & \cellcolor{gray!20}{\textbf{74.36\scriptsize{$\pm$1.78}}} & \cellcolor{gray!20}{\textbf{74.93\scriptsize{$\pm$1.61}}} & \cellcolor{gray!20}{\textbf{75.18\scriptsize{$\pm$1.52}}} & \cellcolor{gray!20}{\textbf{73.66\scriptsize{$\pm$1.55}}} \\

    \midrule

    \multicolumn{10}{c}{\textbf{mTAND}} \\
    \midrule

    ERM & 74.28\scriptsize{$\pm$0.35} & 75.68\scriptsize{$\pm$0.48} & 42.07\scriptsize{$\pm$1.17} & 37.76\scriptsize{$\pm$3.16} & 37.40\scriptsize{$\pm$0.88} & 72.63\scriptsize{$\pm$0.47} & 71.04\scriptsize{$\pm$0.39} & 72.29\scriptsize{$\pm$0.32} & 60.39\scriptsize{$\pm$0.64} \\

    MLDG & 73.46\scriptsize{$\pm$1.54} & 75.54\scriptsize{$\pm$1.71} & 43.61\scriptsize{$\pm$1.72} & 36.82\scriptsize{$\pm$3.11} & 38.11\scriptsize{$\pm$0.95} & 71.84\scriptsize{$\pm$0.96} & 71.43\scriptsize{$\pm$1.08} & 71.96\scriptsize{$\pm$0.96} & 60.35\scriptsize{$\pm$1.21} \\

    ManyDG & \textbf{86.94\scriptsize{$\pm$0.16}} & \textbf{89.52\scriptsize{$\pm$0.22}} & {61.02\scriptsize{$\pm$1.08}} & 40.08\scriptsize{$\pm$2.00} & {41.44\scriptsize{$\pm$1.30}} & \textbf{83.43\scriptsize{$\pm$0.23}} & \textbf{81.84\scriptsize{$\pm$0.17}} & \textbf{82.52\scriptsize{$\pm$0.35}} & {70.85\scriptsize{$\pm$0.34}} \\

    Mixup & 78.00\scriptsize{$\pm$0.96} & 80.39\scriptsize{$\pm$0.91} & 47.07\scriptsize{$\pm$1.05} & 36.68\scriptsize{$\pm$3.16} & 38.77\scriptsize{$\pm$0.61} & 75.71\scriptsize{$\pm$0.69} & 73.62\scriptsize{$\pm$0.68} & 75.33\scriptsize{$\pm$0.59} & 63.20\scriptsize{$\pm$0.77} \\

    MMD & 78.09\scriptsize{$\pm$1.21} & 80.40\scriptsize{$\pm$1.10} & 46.18\scriptsize{$\pm$1.18} & {43.23\scriptsize{$\pm$1.62}} & 38.95\scriptsize{$\pm$0.80} & 74.83\scriptsize{$\pm$0.83} & 73.78\scriptsize{$\pm$0.85} & 74.76\scriptsize{$\pm$0.84} & 63.78\scriptsize{$\pm$0.75} \\

    \cellcolor{gray!20}{\method{}} & \cellcolor{gray!20}{{84.46\scriptsize{$\pm$1.69}}} & \cellcolor{gray!20}{{85.74\scriptsize{$\pm$1.75}}} & \cellcolor{gray!20}{\textbf{67.81\scriptsize{$\pm$1.67}}} & \cellcolor{gray!20}{\textbf{76.38\scriptsize{$\pm$1.71}}} & \cellcolor{gray!20}{\textbf{76.45\scriptsize{$\pm$1.39}}} & \cellcolor{gray!20}{{82.70\scriptsize{$\pm$1.55}}} & \cellcolor{gray!20}{{79.20\scriptsize{$\pm$1.60}}} & \cellcolor{gray!20}{{81.36\scriptsize{$\pm$1.65}}} & \cellcolor{gray!20}{\textbf{79.26\scriptsize{$\pm$1.59}}} \\

    \midrule

    \multicolumn{10}{c}{\textbf{GRU-D}} \\
    \midrule

    ERM & 81.30\scriptsize{$\pm$0.43} & 83.95\scriptsize{$\pm$0.51} & 77.63\scriptsize{$\pm$0.57} & 48.65\scriptsize{$\pm$3.22} & 46.92\scriptsize{$\pm$1.22} & 75.20\scriptsize{$\pm$0.73} & 71.98\scriptsize{$\pm$1.25} & 76.25\scriptsize{$\pm$0.91} & 70.23\scriptsize{$\pm$0.77} \\

    MLDG & 78.36\scriptsize{$\pm$0.93} & 80.48\scriptsize{$\pm$0.98} & 75.05\scriptsize{$\pm$0.96} & 50.22\scriptsize{$\pm$2.65} & 48.16\scriptsize{$\pm$0.82} & 71.97\scriptsize{$\pm$1.14} & 70.37\scriptsize{$\pm$1.85} & 73.47\scriptsize{$\pm$1.75} & 68.51\scriptsize{$\pm$1.04} \\

    ManyDG & 73.74\scriptsize{$\pm$1.78} & 75.82\scriptsize{$\pm$1.95} & 66.76\scriptsize{$\pm$1.80} & 34.55\scriptsize{$\pm$3.97} & 33.67\scriptsize{$\pm$0.98} & 69.27\scriptsize{$\pm$1.75} & 61.90\scriptsize{$\pm$1.04} & 66.90\scriptsize{$\pm$1.26} & 60.33\scriptsize{$\pm$1.55} \\

    Mixup & {83.35\scriptsize{$\pm$0.38}} & {85.59\scriptsize{$\pm$0.40}} & {78.01\scriptsize{$\pm$0.63}} & {50.54\scriptsize{$\pm$2.56}} & {50.30\scriptsize{$\pm$0.64}} & {79.32\scriptsize{$\pm$0.41}} & {76.78\scriptsize{$\pm$0.51}} & {79.95\scriptsize{$\pm$0.32}} & {72.98\scriptsize{$\pm$0.40}} \\

    MMD & 81.09\scriptsize{$\pm$0.60} & 84.33\scriptsize{$\pm$0.50} & 77.35\scriptsize{$\pm$0.33} & 48.13\scriptsize{$\pm$3.10} & 47.85\scriptsize{$\pm$1.06} & 75.67\scriptsize{$\pm$0.47} & 71.74\scriptsize{$\pm$1.37} & 75.67\scriptsize{$\pm$0.88} & 70.23\scriptsize{$\pm$0.73} \\

    \cellcolor{gray!20}{\method{}} & \cellcolor{gray!20}{\textbf{86.84\scriptsize{$\pm$1.57}}} & \cellcolor{gray!20}{\textbf{87.58\scriptsize{$\pm$1.63}}} & \cellcolor{gray!20}{\textbf{81.21\scriptsize{$\pm$1.38}}} & \cellcolor{gray!20}{\textbf{77.34\scriptsize{$\pm$1.92}}} & \cellcolor{gray!20}{\textbf{77.15\scriptsize{$\pm$1.36}}} & \cellcolor{gray!20}{\textbf{84.65\scriptsize{$\pm$1.64}}} & \cellcolor{gray!20}{\textbf{79.85\scriptsize{$\pm$1.64}}} & \cellcolor{gray!20}{\textbf{82.80\scriptsize{$\pm$1.38}}} & \cellcolor{gray!20}{\textbf{82.18\scriptsize{$\pm$1.51}}} \\

    \bottomrule
\end{tabular}}
\end{table*}

\begin{figure*}[!t]
\centering
\includegraphics[width=\textwidth]{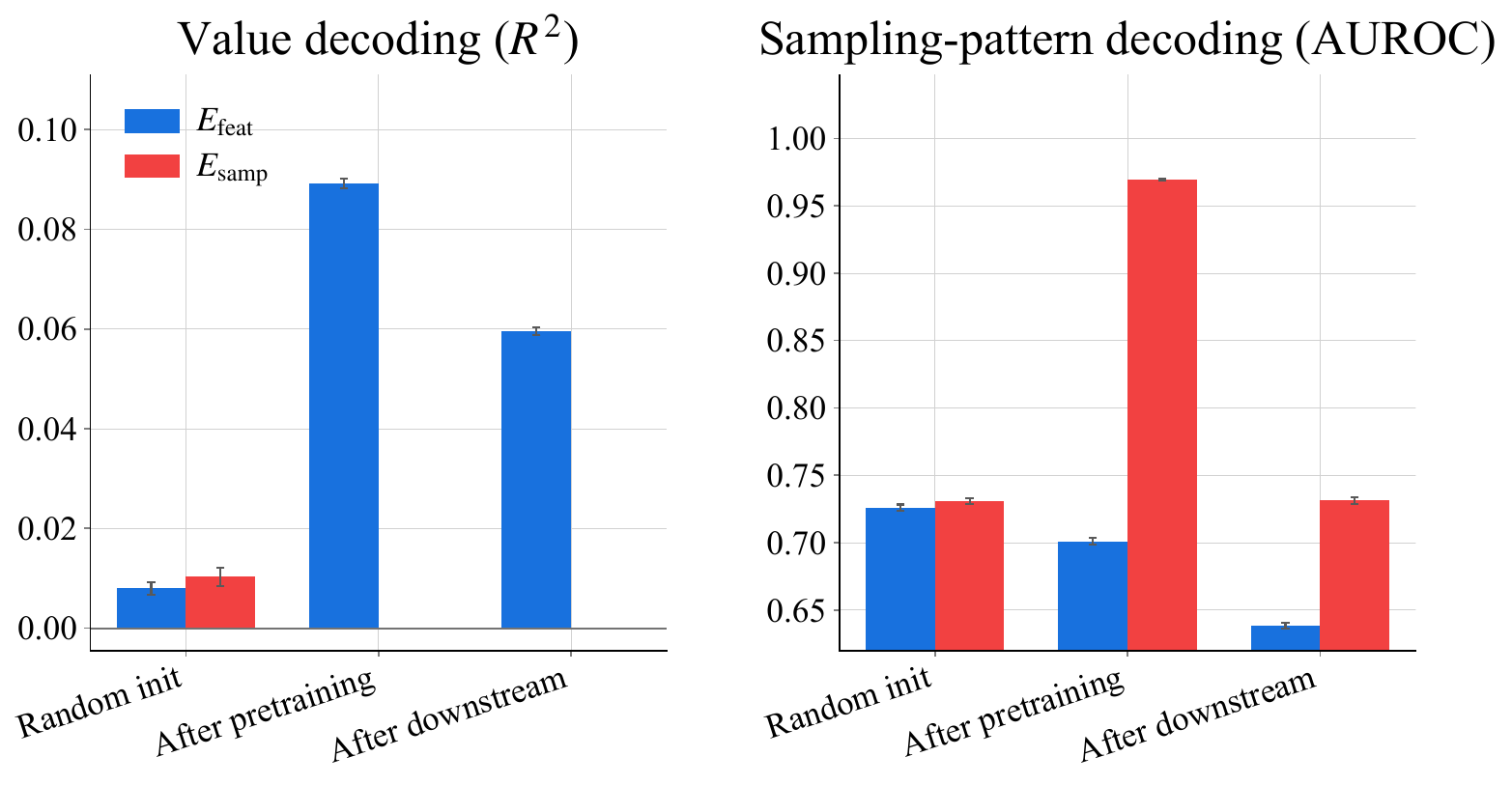}
\vspace{-.15in}
\caption{
Representation analysis of the feature- and sampling-centric encoders.
We measure feature-value information with $R^2$ and sampling-pattern information with AUROC at random initialization, after pretraining, and after downstream training.
}
\label{fig:branch_information}
\vspace{-.15in}
\end{figure*}

\end{document}